\pdfoutput=1
\documentclass[11pt]{article}

\usepackage[final]{acl}

\usepackage{times}
\usepackage{latexsym}

\usepackage[T1]{fontenc}
\usepackage[utf8]{inputenc}

\usepackage{microtype}
\usepackage{float}

\usepackage{inconsolata}

\newcommand\mcrm[1]{{\color{blue!70}#1}}

\newcommand{\gray}[1]{\textcolor{gray}{#1}}
\newcommand{\orange}[1]{\textcolor{orange}{#1}}
\usepackage{graphicx}

\usepackage{adjustbox}
\usepackage{multirow}
\usepackage{amsmath}
\usepackage{comment}
\usepackage{tablefootnote}
\usepackage{hyperref}

\usepackage{soul}

\newcommand\todo[1]{{\textcolor{red}{#1}}}

\newcommand\fbcubed{F-B$^3$}
\newcommand\semcorwsi{SemCor-WSI}

\newcommand\agsilh{AG$_{silh}$}

\title{In the LLM era, Word Sense Induction remains unsolved}%\\ and cheap methods are not dead!}

\author{Anna Mosolova$^{1,2}$, {\bf Marie Candito$^1$} , {\bf Carlos Ramisch$^2$} \\
         $^1$Université Paris Cité, CNRS, LLF, Paris, France \\ 
         $^2$Aix Marseille Univ, CNRS, LIS, Marseille, France \\ \texttt{first.last@u-paris.fr}, \texttt{first.last@lis-lab.fr}}

\begin{document}
\maketitle
\begin{abstract}
In the absence of sense-annotated data, word sense induction (WSI) is a compelling alternative to word sense disambiguation, particularly in low-resource or domain-specific settings.
In this paper, we emphasize methodological problems in current WSI evaluation. We propose an evaluation on a SemCor-derived dataset, respecting the original corpus polysemy and frequency distributions.
We assess pre-trained embeddings and clustering algorithms across parts of speech, and propose and evaluate an LLM-based WSI method for English. We evaluate data augmentation sources (LLM-generated, corpus and lexicon), and semi-supervised scenarios using Wiktionary for data augmentation, must-link constraints, number of clusters per lemma.

We find that no unsupervised method (whether ours or previous) surpasses the strong "one cluster per lemma" heuristic (1cpl). We also show that (i) results and best systems may vary across POS, (ii) LLMs have troubles performing this task, (iii) data augmentation is beneficial and (iv) capitalizing on Wiktionary does help. It surpasses previous SOTA system on our test set by 3.3\%. WSI is not solved, and calls for a better articulation of lexicons and LLMs' lexical semantics capabilities.
\begin{comment}
    
\begin{itemize}
    \item we want to bring attention to WSI task
    \item show that it remains challenging for large language models
    \item highlight some problems of the existing shared tasks (their definition, datasets and metrics)
    \item we propose a solution to this problem by introducing a new dataset
    \item we report the results on it
    \item show how LLMs still can help improve the results on this task indirectly
\end{itemize}
\end{comment}
\end{abstract}

%%%%%%%%%%%%%%%%%%%%%%%%%%%%%%%%%%%%%%%%%%%%%%%%%%%%%%%%%%%%%%%%%%%%%%%%%%%%%%%%%%%%%%%%%%%%%%%%%%%%

\section{Introduction}

Disambiguating the senses of potentially ambiguous words in a text (i.e.~word sense disambiguation, WSD) is a historic NLP task, essential for obtaining a formal representation of a text's meaning. However, this task has the drawbacks of (i) relying on predefined sense inventories of arbitrary granularity and ill-suited for specialized domains, and (ii) requiring labor intensive sense-annotated data, unavailable for most languages of the world. This requirement still holds in the Large Language Models (LLM) era: \citet{sainz-etal-2023-language} and \citet{anonymous2025are} show that open LLMs outperform BERT-based supervised systems only when fine-tuned on sense-labeled data.
%\mcrm{à voir si on descend ça ds le related work, voir aussi \cite{sainz-etal-2023-language} "prompting" de BERT et roBERTa, mais tjrs moins bon que le supervisé. \cite{ortegamartín2023linguisticambiguityanalysischatgpt} étudie capacités de desamb de chatGPT, mais pas d'évaluation sur benchmark classiques et donc pas de comparaison (?)}    

The word sense induction task (WSI) does away with the need for a predefined sense inventory and sense-labeled data (except for evaluation), albeit at the expense of quality. In addition, one of the standard techniques in WSI is to cluster vector representations of a lemma's occurrences. When applied to all content words in a corpus, it provides corpus-dependent pseudo-sense labeling \cite{eyal-etal-2022-large}. Although in the LLMs era the utility of induced senses %-labeled corpora 
is less clear for downstream applications, it remains central in computationally less intensive  environments, %lexical semantics and 
corpus studies such as lexical change detection \citep{schlechtweg-etal-2020-SemEval}, or specific contexts such as scientific literature mining and sense-aware retrieval \citep{eyal-etal-2022-large}.
%is still to consider for computationally less intensive contexts. % NLP applications.  \notes{we could add about domain adaption problem and that not everyone has resources to use llms} (second part done)

Our work focuses on a ``\textbf{full-corpus WSI scenario}'', taking a corpus as input, inducing senses for all content-word lemmas occurring more than once (typically by clustering occurrences), %context-clustering approaches to WSI, which take a corpus as input, cluster all instances of content word lemmas,\footnote{More precisely, all lemmas occurring at least twice, since the hapaxes are trivially clustered into a single cluster.} 
yielding as a by-product pseudo-sense annotations. % (cluster IDs). %This is typically achieved by clustering the lemma corpus instances. 
We note that popular WSI datasets created for SemEval shared tasks \citep{manandhar-etal-2010-SemEval,jurgens-klapaftis-2013-SemEval} %(SemEval2010 Task 14 \cite{manandhar-etal-2010-SemEval}, SemEval2013 Task 13 \cite{jurgens-klapaftis-2013-SemEval}) 
(i) exhibit artificial polysemy levels (because unambiguous lemmas are trivial to process, these datasets over-represent polysemous ones, thus not addressing the task of discovering which lemma is polysemous), (ii) and some exhibit artificial selection of lemmas and number of occurrences per lemma. %SemEval 2013 data also artificially balance the number of occurrences per lemma, regardless of their corpus frequencies. The criteria used to select the covered lemmas  
%may also be arbitrary, but greatly influence evaluation results. 
State-of-the-art WSI systems for English are thus biased towards these unrealistic distributions, and it remains to prove that they work as well when faced with more natural data. 

This paper makes the following contributions:
\begin{itemize}
    %\item We empirically investigate clustering algorithms metrics: building on \citet{Amig2008}, we evaluate whether usual WSI clustering metrics have the required properties, and we empirically compare metrics across datasets (and thus across polysemy levels) and across the target lemmas' parts of speech.
    \vspace{-4pt}
    \item We question the current evaluation in WSI: %setup guiding the development of WSI methods: 
    issues with standard datasets and metrics lead to methodological problems that hinder comparability and reproducibility (\S~\ref{sec:issues}).     
     \vspace{-4pt}
     \item Building upon SemCor, we propose a more natural evaluation framework that %\cite{miller-etal-1993-semantic}, 
    respects the original polysemy and frequency distribution, and benchmark systems across datasets (\S~\ref{sec:variability}). % (and thus polysemy levels).
    %(the WordNet-annotated instances of) SemCor.  
    %\item Given this new framework, we benchmark previous systems, investigate the impact of data augmentation (both extracted from corpora and LLM-generated), and propose and evaluate an LLM-based WSI method (Section~\ref{sec:augmentation}).
    %\item We also investigate a semi-supervised scenario relying on the English Wiktionary as a source for data augmentation, must-link constraints, and number of clusters per lemma (Section~\ref{sec:augmentation}). 
     \vspace{-4pt}
     \item In this new framework, we assess clustering of pre-trained contextualized embeddings across parts of speech, and propose and evaluate direct LLM prompting for WSI for English (\S~\ref{sec:techniques}). 
     \vspace{-4pt}
     \item We evaluate data augmentation sources (LLM-generated, corpus and lexicon), and semi-supervised scenarios using the English Wiktionary for data augmentation, must-link constraints, number of clusters per lemma % as a source for data augmentation, must-link %and \notes{cannot-link} constraints, and number of clusters 
    (\S~\ref{sec:augmentation}). 
\end{itemize}
    
%We find that considering a less artificial evaluation dataset changes the overall picture, with no unsupervised method whether ours or previous) surpassing the strong "one cluster per lemma" heuristic. We also show that (i) results and best systems may vary across POS, (ii) LLMs have troubles performing this task, (iii) data augmentation is beneficial both for semi- and unsupervised systems, and (iv) that capitalizing on a crowd-sourced lexicon such as Wiktionary does help, in particular using must-link constraints and targeting the number of Wiktionary senses as number of clusters.

%\mcrm{Annonce plan?} % Pas trop place... mettre les refs section entre parenthèses avec les contrib?

%%%%%%%%%%%%%%%%%%%%%%%%%%%%%%%%%%%%%%%%%%%%%%%%%%%%%%%%%%%%%%%%%%%%%%%%%%%%%%%%%%%%%%%%%%%%%%%

\section{Related work}
\label{sec:related_work}

We focus here on previous work on WSI compatible with the full-corpus scenario (hence we ignore WSI approaches based on a lexical network such as \citet{panchenko-etal-2017-unsupervised-mean}).
Such corpus-based approaches can be categorized into six groups: 

\paragraph{(i) Clustering contextualized embeddings directly:} the basic technique is to cluster contextualized embeddings produced by masked pre-trained language models \citep{devlin-etal-2019-BERT} (hereafter {\bf PLM}). \citet{lietard-etal-2024-word} perform two-step clustering\footnote{We note the similarity of this method with the semantic frame induction work of \citet{yamada-etal-2021-semantic}.} %\footnote{Note the authors did not mention the similarity with the semantic frame induction work of \citet{yamada-etal-2021-semantic}.} 
using BERT contextualized embeddings of target in-context words. The ``local'' step is a hard clustering of occurrences of a given lemma (the strict WSI task), while the ``global'' step agglomerates the centroids of the local clusters, hence obtaining clusters for ``concepts'' (equivalent to WordNet's synsets). %Interestingly the authors show that the global step helps for strict WSI results. 
Results on SemCor nouns (with more than 10 instances) show that the global step helps for the strict WSI task. The authors report a %very
high non-weighted average {\fbcubed} ($80\%$), but we point out that the average {\fbcubed} is usually \textit{weighted} by each lemma's number of instances, which is all the more crucial in the full-corpus scenario.% \notes{à comparer avec la 1cpl de semcor??}
%\todo{Lietard 1cpl: \\ F-B-Cubed weighted: 72.9 \\
%F-B-Cubed (simple mean): 77.3}
% NB: lietard utilisent le terme "extended bcubed de Amigo, qui est le meme que fuzzyb3 de semeval 2013. le pb du weighing demeure

\paragraph{(ii) Clustering contextualized embeddings enhanced for lexical semantics tasks:} \citet{Abdine_wsi} train a small neural network to maximize the mutual information of pairs of original and perturbed instances.
%TO SHORTEN transform contextualized representations using a small neural network trained using a mutual information %maximization objective. Its goal is %The network is trained 
%to bring the original target word vector closer to its perturbed version.
Then agglomerative clustering (AG) is used on vectors from the hidden layer of the network. AG is performed
%is then used to produce sense vectors, which are clustered using agglomerative clustering 
with either a fixed number of clusters %TO SHORTENas in \citet{amrami2019bettersubstitutionbasedwordsense} 
or dynamically recomputed using a word polysemy quantification score \citep{xypolopoulos-etal-2021-unsupervised}. The proposal of \citet{yava-etal-2024-improving} also falls into this category. It consists in adversarial training of BERT to neutralize morphological and syntactic features, hypothesizing that they introduce noise for lexical semantics tasks. The authors perform K-means clustering of these modified contextualized embeddings, for all SemCor nouns and verbs, excluding those having a unique sense, and senses with less than 10 occurrences.%occurring more than once and having more than 10 occurrences. 
%\todo{technically, they excluded all that have less than 10 occs $\rightarrow$ used only those with more than 9 occurrences, but i don't think it's really important, so I'd leave like this}

There are other works enhancing contextualized embeddings for lexical tasks, but not evaluating them for WSI. The main evaluation task for these is the Word-In-Context task (WiC, \cite{pilehvar_wic_2019}), a binary classification task to decide whether two instances of the same lemma correspond to the same sense or not. In this vein, we can cite {{\bf MirrorWiC} \cite{liu-etal-2021-MirrorWiC}  and the model of \citet{mosolova-etal-2024-injecting} (which we will dub as {\bf BERT-Wikt}). Both models were fine-tuned using contrastive learning, which teaches the model to bring semantically similar examples closer, while pushing dissimilar ones apart. MirrorWiC leverages self-supervised contrastive learning by using similar examples created by automatically alternating the original phrase. BERT-Wikt employs semi-supervised contrastive learning by using exemplars of the same sense from Wiktionary as similar ones.

\paragraph{(iii) Clustering vectors of BERT substitutes:} In the Language-model Substitution with Dynamic Patterns model (\textbf{LSDP}), \citet{amrami2019bettersubstitutionbasedwordsense} build vectors of BERT substitutes for each target instance, then cluster these vectors using agglomerative clustering. Hearst-like symmetric patterns are used to improve the quality of substitutes. %TO SHORTEN To address the challenge of defining the appropriate number of senses for a given lemma, they use a two-step strategy, first inducing 10 senses for each lemma, classifying these as weak or strong, and merging each weak one with its closest strong one.
%They also explore usage of dynamic patterns to inject target word which improves substitutes quality and strong and weak clusters to dynamically define the number of clusters for each lemma. 
\citet{eyal-etal-2022-large} focus on scaling up this substitute-based technique, so that it can be used in the full-corpus scenario%TO SHORTEN (which they call ``all-words WSI'')
. The authors induce the senses of the 16k single-token words of the BERT-large (whole-word masking) vocabulary, and obtain a sense-labeled version of the English Wikipedia.
 
\paragraph{(iv) Learning sense embeddings using a masked language modeling objective:} \citet{ansell-etal-2021-polylm} propose the {\bf PolyLM model}, which learns contextualized sense embeddings using a language modeling objective. For each lemma in the vocabulary, the model learns a fixed number of sense representations, and assigns in-context probabilities for each sense.
%, the realization and probability of which depends on the context. 
The model builds on the assumptions that the probability of a word in context is the sum of the probabilities of all its senses and that for a given word occurrence, one of its senses should be more plausible than all the other ones. %The model is a combination of two small adapted BERT variants trained from scratch using a language modeling loss, and two losses favoring the above assumptions.
As a by-product, the model produces a probability distribution over senses for each word in context, which can be used to perform WSI. The authors report the SOTA results on SemEval 2010 and SemEval 2013 datasets.

\paragraph{(v) Latent-variable models:}  \citet{Amplayo_Hwang_Song_2019} employ a latent variable model that models senses as distributions over multiple topics and uses target-neighbor pairs to induce more fine-grained senses and filter out the irrelevant ones.

%\subsection{WSI algorithms (corpus instances' clustering (not graph-based))}
\begin{comment}
\begin{itemize}
\item \gray{amrami \cite{amrami2019bettersubstitutionbasedwordsense} dynamic patterns with vectors from substitutes, dynamical number of clusters through classifying them into weak and strong ones (didn't improve the result)}
\item \gray{polylm combination of two BERT models trained from scratch using special loss functions, this model produces 7 sense embeddings for each token with their probability for the context}
\item \cite{eyal-etal-2022-large} Louvain community detection algorithm on a graph of BERT substitutes 
\item \gray{abdine MIM model \cite{Abdine_wsi}}
\item \citep{yava-etal-2024-improving}: K-means + BERT with adversarial training to make it forget morpho, tested on a part of semcor with only polysemous nouns and verbs having at least 10 examples
    \item \citet{panchenko-etal-2017-unsupervised-mean}
\end{itemize}
\end{comment}

%% formulation de Liétard: From lists of substitutes, they build a graph of substitutes in which they find communities and then assign each occurrence to a community of substitutes to find the wordsenses

%\citet{arefyev-etal-2020-always} continue with the substitutes approach, but explore several PLMs and more sophisticated target word injection methods. They also use agglomerative clustering and compute number of clusters by maximizing the silhouette score for each lemma. 

%specially designed loss functions.

\paragraph{(vi) WSI using LLMs:} Larger decoder-only models have also been evaluated on lexical semantics tasks. Some have been shown to perform well for the WiC task \citep{hayashi-2025-evaluating}. \citet{ortegamartín2023linguisticambiguityanalysischatgpt} report good ability of ``ChatGPT'' to identify ambiguity for specific words. %LLMs were also tested on the WSD task.
\citet{sumanathilaka2024llmsassistambiguityquantitative} investigate LLMs' capabilities for WSD, using the English Wiktionary-derived FEWS dataset \cite{blevins-etal-2021-fews}. Results on a subset of the FEWS test set seem high but are unfortunately not compared to previous works. \citet{sainz-etal-2023-language} and \citet{anonymous2025are} show that open LLMs do outperform BERT-based supervised systems, but only when fine-tuned on sense-labeled data.
Coming back to the WSI task, we can cite only one work involving LLMs: \citet{periti-etal-2024-automatically} fine-tune LLMs on lexicographic definitions and exemplar sentences, for these LLMs to generate a definition given a word in context. 
%The WSI is performed by clustering the embeddings of generated definitions and evaluated on the lexical semantic change dataset \cite{schlechtweg-etal-2020-SemEval}.  
The authors then perform WSI by clustering the embeddings of generated definitions, evaluated on a lexical semantic change dataset.% \cite{schlechtweg-etal-2020-SemEval}. 

Importantly, these LLM approaches to WSD and WSI are computationally intensive, prompting the LLM once for each instance to disambiguate. 
%TO SHORTEN of direct prompting of LLMs for WSI. We propose a more lightweight %less energy intensive 
%technique, asking the LLM to cluster sets of instances of a given lemma.

%TODO: expliquer qu'on traite un modèle de chaque catégorie In this paper, we evaluate at least one representative model from each category. This section presents results for groups (i) and (iv); results for groups (ii) and (iii) are discussed in \S \ref{sec:techniques}, along with an extension proposed in \S \ref{sec:augmentation},
In the following, (i) to the best of our knowledge, we report the first results of directly asking the LLM to cluster sets of instances of a given lemma, a much more lightweight technique; (ii) we compare WSI performance across datasets, target lemma POS, and evaluation metrics for models falling into category (iii) (LSDP) and (iv) (PolyLM). The approaches (i) and (ii) are the focus of \S \ref{sec:techniques}.
%%%%%%%%%%%%%%%%%%%%%%%%%%%%%%%%%%%%%%%%%%%%%%%%%%%%%%%%%%%%%%%%%%%%%%%%%%%%%%%%%%%%%%%%%%%%%%%

%\section{Methodological problems of the existing WSI shared tasks}
\section{Issues in WSI evaluation}
\label{sec:issues}

% In this section, we detail why the datasets and evaluation methodology of previous WSI shared tasks do not match the scenario of inducing senses for all lemmas of a given corpus.

% Summary: extremely heterogeneous and low reproducibility landscape :
% task definition and terminology (sense, concept, synset, and related frame) induction, link to wsd?
% data: target POS, lemma and instance selection in standard (SemEval) and new datasets (letard,conLL25),
% metrics: different choices, often with complex properties (e.g. NMI), details in their application (e.g. average over instances/lemmas, weighted by POS), graded vs. hard assignment, 
% setting: tuning hyperparams, %absence of simple baselines such as 1 cluster per lemma/instance.
% $\implies$ comparing papers and reproducing/replicating results is very hard/impossible.

The evaluation of WSI models relies on datasets containing manual sense assignments for instances of a set of lemmas, and on metrics assessing how well the automatic assignment matches the manual one. Datasets have issues related to lemma and instance selection and dev/test data splits. They are often associated to heterogeneous and incomplete metrics, resulting in a complex landscape in which it is extremely difficult to compare, reproduce and/or replicate results. %Thus, models evaluated on them present methodological weaknesses with the use of test data to tune the systems and omitting competitive trivial baselines. \notes{refer for details to the chapter in the appendix} 

\subsection{Datasets}
Two popular datasets for the evaluation of English WSI are those of SemEval 2010 Task 14 \citep{manandhar-etal-2010-SemEval} and SemEval 2013 Task 13 \citep{jurgens-klapaftis-2013-SemEval}, used e.g. by \citet{amrami2019bettersubstitutionbasedwordsense,Abdine_wsi,eyal-etal-2022-large} (referred as SE10 and SE13 hereafter). See Appendix~\ref{app:wsidatasets} for details on these and other datasets.
% trott-bergen-2021-raw - 112 polysemous words, ~600 sentence pairs, graded similarity
%As they were used in shared tasks, they are associated with their respective evaluation metrics.
% such as one cluster per lemma/instance. 

%\todo{TABLE WITH ALL DATASETS COMPARISON}

%\notes{existing datasets do not satisfy the condition of the natural distribution of senses, instances, lemmas, which influences sense distribution making it unnatural}

%\paragraph{Lemma selection and instance selection}: over-representation of polysemy (lemmas) + non-natural distribution of instances (merge below)

\paragraph{Pre-defined sense inventory evaluation bias} Most WSI works evaluate systems using gold data labeled with senses from a pre-defined inventory. This introduces a bias since the granularity of sense distinction may vary across lexical resources and target objective tasks. \citet{herman-jakubicek-2024-shadowsense} proposed an evaluation dataset for Czech and English, later extended to 6 languages for the upcoming CoNLL 2025 shared task on robust WSI\footnote{{\scriptsize \url{https://projects.sketchengine.eu/conll2025}}}, designed to address this specific bias. 

\paragraph{Senses distribution} Both SE10 and SE13 datasets, as well as CoNLL 2025 robust WSI dataset, tend to over-represent polysemous and frequent lemmas. The first source of bias lies in the selection of lemmas, whose criteria are made explicit in neither of the datasets. Once lemmas are selected, the instances included in the test set come from OntoNotes v1.0 (SE10) and from the Open American National Corpus (SE13), but again the selection of these instances is not motivated. In SE13, all lemmas have at least 22 instances, with the majority having exactly 100. The CoNLL-25 WSI dataset focuses on 25 polysemous lemmas per language. As a result, WSI evaluation tends to ignore monosemous lemmas, albeit their high corpus frequency. Although polysemy detection models do exist, they are never applied in the context of WSI  \citep{springorum-etal-2013-detecting,lossio-ventura-etal-2016-automatic,habibi-etal-2021-homonymy}. In short, there is a mismatch between the sense distribution in these popular datasets and the more realistic full-corpus WSI scenario.

%Polysemy and homonymy detection using translations \citep{habibi-etal-2021-homonymy}, for specialised language \citep{lossio-ventura-etal-2016-automatic}, or for function words \citep{springorum-etal-2013-detecting}.

%\todo{where all lemmas have at least 2 senses and more than one example or even an equal number of examples of each sense which is an extremely rare situation in the real corpora}.
\paragraph{Data splits} While SE10 does not provide development set at all,\footnote{SE10's ``training'' %\notes{this is training set} 
data contains no annotation.} SE13 provides a trial dataset whose senses distribution is extremely different from the test set (different annotators, data sources, number of instances and polysemy levels). Participants who optimized their systems on this trial dataset obtained lower scores in the campaign \citep{jurgens-klapaftis-2013-SemEval}. More recent WSI work ignores the trial data, leading to a problematic use of the test set for hyperparameter tuning and comparison of configurations %, and/or ablation studies 
\citep{amrami2019bettersubstitutionbasedwordsense,Abdine_wsi}.
%\citep{amrami2019bettersubstitutionbasedwordsense,ansell-etal-2021-polylm,Abdine_wsi}.
%using it (ablation studies by Abdine, Amrami and Ansell are made using test set). 

%\paragraph{Task definition} SemEval 2010 addressed traditional WSI (hard clustering), whereas SemEval 2013 focused on graded WSI (soft clustering). Subsequent work proposed variants like concept induction \citep{lietard-etal-2024-word} and multi-annotator WSI \citep{herman-jakubicek-2024-shadowsense}. It is not obvious that systems designed for these tasks are useful in the full-corpus scenario, or whether results generalize across tasks, especially given the heterogeneity of associated datasets and evaluation metrics. 

%and it is not obvious that systems performing best on this task will perform equally well on other corpora with more naturally distributed senses. \notes{the problem here is not senses distribution, but uselessness of this task in real life? maybe delete this point}

%%%%%%%%%%%%%%%%%%%%%%%%%%%%%%%%%%%%%%%%%%%%%%%%%%%%%%%%%%%%%%%%%%%%%%%%%%%%%%%%%%%%%%%%%%%%%%%

\subsection{Metrics}
%\label{subsec:metrics}

Clustering is notoriously difficult to evaluate, with different metrics capturing different properties. In addition, most metrics are sensitive to sense distribution, questioning cross-dataset replicability. % in the full-corpus scenario.

\paragraph{Metrics heterogeneity} Different evaluation metrics were used in SE10 (V-measure and Paired F-score) and SE13 (Fuzzy NMI and Fuzzy {\fbcubed}\footnote{The ``fuzzy'' versions are needed as SE13 asks soft clustering. Nonetheless, the SE10 metrics could have been adapted for soft clustering.}). The upcoming CoNLL 2025 shared task on robust WSI uses yet another metric based on rand index to take into account multiple gold annotations %for a given instance
\citep{herman-jakubicek-2024-shadowsense}.
While previous work reporting results on SE datasets use the corresponding shared task metrics, works evaluated on other datasets use different metrics, making comparison all the more difficult. For instance, \citet{yavas-2024-assessing} use adjusted rand index,  \citet{lietard-etal-2024-word} use {\fbcubed}, \citet{periti-etal-2024-automatically} use rand index, adjusted rand index, and purity, and
%Typically, to evaluate the performance of the clustering systems, the authors reuse the metrics used in the corresponding SemEval tasks. In previous works, it was already noted that the proposed metrics have some problems. 
\citet{komninos-manandhar-2016-structured}, following \citet{li-etal-2014-improved}, adapt V-measure to use the best-upper-bound entropy estimator instead of maximum likelihood to alleviate some of its problems.

%Although \citet{amrami2019bettersubstitutionbasedwordsense} use the SE metrics, they emphasize that small senses should have as much weight as larger ones, proposing to compare the number of senses/clusters using Spearman correlation. In our paper, we rather advocate for metrics computed using instance-based weighting, more suitable for full-corpus WSI.

\paragraph{Metric properties} 
%More precisely, we retain the {\fbcubed} metric (F-score of the $B^{3}$ precision and recall \cite{bagga-baldwin-1998-entity}, a choice based on the extensive analysis of clustering metrics \citet{Amig2008}. The authors 
\citet{Amig2008} test the sensibility of metrics to four desirable properties: \textit{\textbf{H}: cluster homogeneity} (clusters should not mix items belonging to different classes), \textit{\textbf{C}: completeness} (items belonging to the same class should be grouped in the same cluster), \textit{\textbf{RB}: rag bag} (adding an example of another class into clean cluster is worse than adding it into a mixed one) and \textit{\textbf{SQ}: clusters size vs. quantity} (a small error in a big cluster should be preferable to a large number of small errors in small clusters). As shown in Table \ref{tab:amigo_new_metrics_short}, {\fbcubed} is the only one sensitive to all four properties.

\begin{table}[!ht]
    \centering
    \begin{tabular}{c|c|c|c|c}
        \textbf{Metric} & \textbf{H} & \textbf{C} & \textbf{RB} & \textbf{SQ} \\
        \hline
        Rand index & $\surd$ & $\surd$ & $\times$ & $\times$ \\
        Paired F-score & $\surd$ & $\surd$ & $\times$ & $\times$ \\
        \hline
        NMI & $\surd$ & $\times$ & $\times$ & $\surd$ \\
        V-measure & $\surd$ & $\surd$ & $\times$ & $\surd$  \\
        \hline
        B$^3$ Precision & $\surd$ & $\times$ & $\surd$ & $\times$  \\
        B$^3$ Recall & $\times$ & $\surd$ & $\times$ & $\surd$ \\
        {\fbcubed} & $\surd$ & $\surd$ & $\surd$ & $\surd$ \\
    \end{tabular}
    \caption{Sensibility of clustering metrics to the properties defined by \citet{Amig2008}. Table \ref{tab:amigo_new_metrics} in Appendix \ref{app:metrics-properties} illustrates these properties on use cases.}
    \label{tab:amigo_new_metrics_short}
\end{table}

%\car{Work in progress paragraph below}

\paragraph{Metric combination and comparison} 

%The {\fbcubed} metric is calculated for each lemma. To better account for the full corpus WSI scenario, the average {\fbcubed} over all target lemmas must be weighted by each lemma's number of instances, but this not always clearly stated \citep{lietard-etal-2024-word}.\footnote{This was required for SemEval 2010 since the number of instances varied across lemmas. For SemEval 2013, it did not vary, so weighted average and plain average are equivalent.}
%\notes{SemEval 2013 doesn't use it, it's the papers afterwards who started doing it to show the mean result, because NMI is always awful} 
%Moreover, 
Works using the SE13 data report the geometric mean of fuzzy {\fbcubed} and fuzzy NMI \citep{amrami2019bettersubstitutionbasedwordsense,ansell-etal-2021-polylm}. However, the latter is insensitive to completeness, artificially increasing when a class is divided into homogeneous clusters. Moreover, statistical significance is never reported, which weakens systems' comparison. Powerful significance tests for \fbcubed{} are computationally intensive because they require re-building the clusters for each bootstrapped resample. In our work, we test statistical significance for part of our experiments only (in \S \ref{sec:techniques}). %It is possible that, given the dataset sizes and effect magnitude, some models actually do not overcome simple baselines such as one-cluster-per-lemma, competitive in the full-corpus scenario (\S~\ref{sec:variability}).%\footnote{Significance is tricky to address for clustering metrics, but this is not a reason to completely overlook it.}

For full-corpus WSI, it is crucial to use a metric counting instances (like {\fbcubed}), compared to pair-based metrics, which overuse large gold classes, or compared to cluster-based metrics like NMI, which use proportions of gold classes in clusters, independently of their sizes. For these reasons, we adopt weighted average {\fbcubed} in our experiments (except in \S~\ref{sec:variability}, on cross-dataset variability).

\section{Variation of systems' performance across datasets, metrics and POS}
%\section{Variability in dataset, metrics and POS}
\label{sec:variability}

\begin{comment}
\begin{itemize}
    \item introduce dataset with more natural polysemy and frequency distribution
\item compare the best previous systems on this dataset and existing popular ones (SemEval 2010 and 2013) 
\item compare them to LLM-based WSI clustering
\item and empirically investigate whether the rank of these systems are stable across datasets, metrics and POS of the lemma
\end{itemize}
\end{comment}

%\notes{we propose a new evaluation framework based on semcor. we choose truly random lemmas with all their instances (unlike SemEval 2010 2013)}

In this section, we investigate the impact of datasets' characteristics (polysemy level, target lemmas POS) and of evaluation metrics on performance in WSI.

We first describe our WSI evaluation dataset %\textbf{{\semcorwsi}} 
extracted from SemCor \citep{miller-etal-1993-semantic}, %. By randomly selecting target lemmas, we obtain sets that
respecting the original distributions of %number of 
corpus occurrences and %number of 
senses.

We then compare WSI performance across (i) recent state-of-the art systems (LSDP and PolyLM)\footnote{We focused on the WSI previous works for which the code is available and functional, or for which there are reported results on Semeval 2010 and/or 2013 WSI datasets.}, and (ii) direct LLM prompting for WSI. %(recent previous systems plus direct prompting of LLMs) % for WSI), 
%across datasets with varying polysemy levels, across target lemma POS (verbs, nouns, adjectives), and across usual WSI clustering metrics, in order to study the impact of these traits on performance.%the systems ranking.
% rephrased TO SHORTEN the objective being to study how metrics and target lemma frequency, polysemy level and POS impact the systems' ranking. %as well as previously mentioned SemEval 2010 and SemEval 2013 datasets. Additionally, for all datasets, we will report the results of several large language models using direct prompting for the target task. 
 
%\notes{extract lemmas with natural distribution of senses in a corpus, so semcor is a natural choice. later we'll show that this dataset shows better how the system will perform in real settings}

%%%%%%%%%%%%%%%%%%%%%%%%%%%%%%%%%%%%%%%%%%%%%%%

\subsection{{\semcorwsi}: extraction from SemCor}

To overcome the flaws of the usual WSI datasets, we propose to tune and evaluate WSI systems on a dev and test sets extracted from SemCor 3.0 \citep{miller-etal-1993-semantic}, a WordNet sense-annotated corpus\footnote{We used the brown1 and brown2 parts, which are the only ones having sense annotations for all open class words.}.
%SemCor consists of 352 English texts from Brown Corpus manually annotated using WordNet 1.6 senses. It was previously used for WSD tasks \todo{cite something (and WSI too..\footnote{\notes{to the best of my knowledge, no one published their partition, so we can't reuse}})}. SemCor 3.0 is composed of three parts: \textit{brown1} (103 files), \textit{brown2} (83 files) and \textit{brownv} (166 files), where only two first parts have sense annotations for all open class words. Consequently, we use only these two parts to compile the new dataset. 
%We include single- and multiword lemmas in the dataset. 
We extract three sub-datasets of sense-annotated instances of verbs, nouns and adjectives%\footnote{\notes{TODO: find a reason why we did not use adverbs}}
: for each POS, we first consider the full lexicon of SemCor lemmas having this POS (including both single- and multiword lemmas) and occurring at least twice. Then for each POS, we (i) %iteratively
randomly select lemmas until we obtain approximately 10,000 corpus instances, and (ii) randomly split each POS's dev and test sets, keeping disjoint sets of lemmas, targeting the same number of instances, lemmas, and polysemy levels in both parts. We kept the first sense for instances annotated with multiple senses. So overall, the dataset contains both monosemous and polysemous lemmas, with instance counts per lemma varying from 2 to several hundreds, with a similar polysemy level in dev, test and full SemCor, as shown in Table \ref{tab:semcor_statistics} (Appendix \ref{app:semcorwsi_statistics})\footnote{The code and the dataset are available at: {\scriptsize \url{https://github.com/anya-bel/fullcorpus_wsi}}}. %We believe that a system demonstrating advanced performance on this dataset is expected to show comparable results on new, previously untested corpora.

Note that while this scenario addresses the unnaturalness of the SE10 and 2013 dataset pointed in \S \ref{sec:issues}, it does not address the pre-defined sense inventory evaluation bias (addressed by \citet{herman-jakubicek-2024-shadowsense} and in the upcoming CoNLL 2025 shared task on robust WSI). We leave it for future work to combine a scenario evaluating both on a natural sense distribution and circumventing the pre-defined senses' bias.

\subsection{Experimental protocol}

We evaluate five models, starting with the state-of-the-art PolyLM model \citep{ansell-etal-2021-polylm}.%\footnote{{\scriptsize \url{https://github.com/AlanAnsell/PolyLM}}}
We compare PolyLM base (54M parameters) and PolyLM large (90M parameters). Another model is LSDP by \citet{amrami2019bettersubstitutionbasedwordsense}\footnote{{\scriptsize \url{https://github.com/asafamr/BERTwsi}}}. We report the average and standard deviation of 10 runs as per the authors' methodology. To adapt LSDP for the {\semcorwsi} dataset, we modified the process of determining strong and weak senses (see Appendix \ref{app:amrami} for details). 
%For both models, 
We reuse the default hyperparameters set by the authors.

%%%%%%%%%%%%%%%%%%%%%%%%%%%%%%%%%%%%%%%%%%%%%%%

\begin{table*}[h!]
    \centering
    \begin{adjustbox}{max width=\textwidth}
    \begin{tabular}{c|cc|cccc|cc}
        \hline
        \textbf{Model} & \multicolumn{2}{c|}{\textbf{SemEval 2013}} & \multicolumn{4}{c|}{\textbf{SemEval 2010}} & \multicolumn{2}{c}{\textbf{\semcorwsi}} \\
        & \textbf{Fuzzy-NMI} & \textbf{Fuzzy-{\fbcubed}} & \textbf{V-M} & \textbf{Paired F-S} & \textbf{NMI} & \textbf{{\fbcubed}} & \textbf{{\fbcubed}} & \textbf{NMI} \\
        \hline
        PolyLM large & \textbf{23.7} & \textbf{66.7} & \textbf{43.6} & 67.5 & 6.2 & 49.2 & 73.0 & \textbf{33.6} \\
        PolyLM base & 23.0 & 65.4 & 41.8 & 66.4 & 6.2 & 49.1 & 71.3 & 31.5 \\
        LSDP & 21.1[$\pm$0.6] & 64.1[$\pm$0.5] & 38.9[$\pm$1.0] & \textbf{70.7}[$\pm$0.4] & 4.6[$\pm$0.1] & 52.8[$\pm$0.2] & 71.0[$\pm$0.4] & 32.1[$\pm$0.7] \\
        %\todo{abdine(their)} & 22.74[$\pm$0.5] & 62.8[$\pm$0.49] & 44.83[$\pm$1.08] & 67.74[$\pm$0.78] & & & & \\
        \hline 
        Llama 3.1 8B & 2.3[$\pm$0.4] & 57.1[$\pm$0.5] & 16.5[$\pm$0.9] & 49.3[$\pm$1.2] & 7.3[$\pm$0.5] & 49.6[$\pm$0.9] & 59.7[$\pm$1.0] & 19.4[$\pm$0.8] \\
        Llama 3.3 70B & 8.9[$\pm$0.4] & 44.2[$\pm$1.8] & 29.4[$\pm$0.9] & 49.7[$\pm$4.8] & 8.1[$\pm$0.6] & 49.6[$\pm$1.6] & 64.2[$\pm$0.9] & 27.8[$\pm$1.1] \\
        GPT-4o & 16.9[$\pm$0.5] & 58.6[$\pm$1.6] & 36.3[$\pm$2.0] & 63.9[$\pm$2.0] & 7.1[$\pm$0.3] & 47.7[$\pm$1.9] & 66.9[$\pm$0.7] & 29.2[$\pm$1.2] \\
        %o1  \\
        \hline
        1cpl & 0.0 & 61.23 & 0.0 & 63.5 & 0.0 & \textbf{64.1} & \textbf{73.6} & 28.1 \\
        1cpex & 6.9 & NA & 31.7 & 0 & \textbf{19.5} & 8.0 & 24.1 & 20.7 \\
        \hline
        %Scores Sp. corr & %\multicolumn{2}{|c|}{0.94} & \multicolumn{2}{c}{0.82} & & \\
        %\hline
    \end{tabular}
    \end{adjustbox}
    \caption{WSI results on 3 datasets. PolyLM large/base: for SE10/SE13, results reproduced using models of {\scriptsize \url{https://github.com/AlanAnsell/PolyLM}}, for \semcorwsi{}, results obtained using their code. LSPD: substitutes obtained with BERT-large. For SE10/SE13, results reproduced with the code of \citet{amrami2019bettersubstitutionbasedwordsense}, for \semcorwsi{}, results obtained using the adapted code (see text).}
    \label{tab:results_on_existing_datasets}
\end{table*}

We also test 3 large language models: the proprietary GPT 4-o \citep{openai2024gpt4technicalreport} and two open-source models: Llama 3.1 8B Instruct \cite{grattafiori2024llama3herdmodels}  and Llama 3.3 70B Instruct (4 bit). For each model, we use identical prompts adapted to the specific task (full prompts, prompt tuning details and exact LLMs versions are given in Appendix \ref{app:prompts}). We report both the average and standard deviation of five runs for each LLM.

For each dataset, we provide two simple baselines: one cluster per lemma (1cpl) and one cluster per example (1cpex).

\subsection{Results and discussion}
\label{sec:all_datasets_results}

Table~\ref{tab:results_on_existing_datasets} shows the performance of each model (results per POS provided in Appendix \ref{app:perposresults}). For SE10, we provide also the non-fuzzy versions of the SE13 metrics, and for comparison, we use these for {\semcorwsi} ({\fbcubed} and NMI). Globally, PolyLM-large is the best-performing model having the highest results for 4 metrics across three datasets (but this does not hold for all individual POS). \textbf{None of the models surpasses the high 1cpl baseline in {\fbcubed} on {\semcorwsi}}, a result likely to hold for full-corpus scenario on other corpora. %suggesting that most models are likely to under-perform compared to the 1cpl solution in real-world corpora. The high 1cpl baseline for {\semcorwsi} stems from the lower polysemy level in the full-corpus scenario.  %\todo{concerning per part of speech performance,  les meilleurs modèles varient-ils selon la POS? => un peu, mais polylm large domine} \notes{\textbf{verbs}: polylm large best 3 times, polylm base 1 time, Llama 70B twice;\textbf{nouns}: polylm large 3 times, LSDP 2 times, 1cpv twice, 1cpex once; \textbf{adjectives}: polylm base once, LSDP once, 1cpv twice}. . 

%\todo{results on llm change a lot (std)}

The evaluation metrics do not always follow the same patterns.  %For instance, when evaluating SemEval 2010 with the non-fuzzy versions of the SemEval 2013 metrics, the top performers are the baselines: 1cpl for {\fbcubed} and 1cpex for NMI. 
On SE10, the best model differs for each of the four reported metrics. For SE13 and \semcorwsi, {\fbcubed} and NMI (and their fuzzy versions) correlate better.  %The weak Spearman correlation between Paired F-Score and {\fbcubed} (\textit{0.12}) and between V-measure and {\fbcubed} (\textit{-0.29}) suggests that the models might have been overly optimized for specific metrics. 

Moreover, on {\fbcubed}, the 1cpl baseline performs best for SE10 and {\semcorwsi}. For SE10, it suggests that optimizing the systems for the original metrics is detrimental when changing metrics.
%Additionally, {\fbcubed} is the only metric for which neither 1cpex nor 1cpl score zero, illustrating a balance between these aspects.\car{1cpex has FB3=8 not far from zero, whereas NMI is not zero for both baselines on Semcor-WSI. I suggest removing this sentence} \notes{good point, it will save us 3 lines, @Marie, if you see this and agree, comment it}

Performance of LLMs is globally much lower (except on verbs, see Table \ref{tab:results_on_existing_dataset} in Appendix), with sometimes huge variance. Llama models struggle with processing lemmas with large numbers of examples (present in SE10, and in the full-corpus scenario). They tend to forget the task after 300 examples and either halt generation, repeat the same answer, or produce irrelevant sentences\footnote{Processing sets of instances of a given lemma in batches should be investigated, but requires to merge the sense inventory induced for each batch.}. While the GPT-4o model does not suffer from this limitation, all LLMs occasionally produce unnecessary explanations, ask for a sense inventory, or simply "refuse" to perform the task. Moreover, parsing answers revealed difficult, in particular  
%the answer-parsing step complicates the application of these models, in particular, 
for sense applicability degrees in SE13.% where each sense should be accompanied by its applicability degree. 
Note though that despite these flaws, LLMs sometimes produced interpretable cluster names, %. At the moment, the results produced by the traditional clustering methods seem to be more reliable than the ones generated by the tested LLMs, despite the fact that LLMs can sometimes produce meaningful cluster names, 
which is not straightforward % was: impossible
with traditional approaches.

%%%%%%%%%%%%%%%%%%%%%%%%%%%%%%%%%%%%%%%%%%%%%%%%%%%%%%%%%%%%%%%%%%%%%%%%%%%%%%%%%%%%%%%%%%%%%%

\section{Investigating full-corpus WSI}
%\section{Full-corpus WSI techniques}
\label{sec:techniques}

We reported earlier
%in previous section 
that the top-performing systems do not surpass the 1cpl baseline when switching to a more naturally distributed dataset. This suggests that over-representation of polysemy in earlier datasets may have influenced the systems' design. 
In this section, we investigate the performance achievable on {\semcorwsi} using the basic technique of clustering the contextualized embeddings of a given lemma instances and we hypothesize that performance may vary depending on polysemy injecting lexical information either using unlabeled . We perform a grid search using (i) two clustering algorithms which automatically determine the number of clusters (X-Means and \agsilh), (ii) two BERT PLMs, plus PLMs fine-tuned to better perform on WiC task. This allows us to assess which model and algorithm combination performs best on a more naturally distributed dataset. 

%%%%%%%%%%%%%%%%%%%%%%%%%%%%%%%%%%%%%%%%%%%%%%%

\subsection{Experimental protocol}

\begin{table*}[!ht]
    \centering
    \begin{tabular}{ll|lll|ll}
        \hline
        \textbf{Model} & \textbf{Algo} & \textbf{Verb} & \textbf{Adj} & \textbf{Noun} & \textbf{All POS} & \textbf{$_w$Avg} \\
        \hline
        \hline
        \multicolumn{7}{l}{\textbf{Unsupervised}} \\
        %\hline
        \multirow{2}{*}{BERT-b-u} & \agsilh & 64.8 & 75.6 & 72.3 & 70.6 & 70.8\\
        & X-Means & 62.9[±0.1] & \textbf{76.5}[±2.2] & 73.7[±0.3] & 70.6[±0.8] & 71.1 \\
        \hline
        \multirow{2}{*}{BERT-l-u} & \agsilh  & \textbf{65.8} & 75.7 & 72.3 & \textbf{71.1} & 71.1 \\
        & X-Means & 63.2[±0.6] & 75.5[±1.2] & \textbf{\textcolor{blue}{74.8}}[±0.1] & 70.2[±0.1] & \textbf{71.5} \\
        %\hline
        %\multirow{2}{*}{ModernBERT-base} & \agsilh & 63.5 & 74.8 & 65.5 & 67.1 & 66.9 \\
        %& X-Means & 61.3[±1.6] & \textbf{\textcolor{blue}{78.7}}[±0.0] & 72.1[±0.8] & 70.2[±0.4] & 70.3 \\
        \hline
        %\multirow{2}{*}{ModernBERT-large} & \agsilh & 63.1 & 74.2 & 65.8 & 67.3 & 66.8 \\
        %& X-Means & \textbf{71.1}[±0.5] & \textbf{\textcolor{blue}{78.7}}[±0.0] & 72.7[±0.4] & 62.1[±0.9] & 70.9 \\
        %\hline
        \hline
        \multicolumn{7}{l}{\textbf{Self-supervised}} \\
        %\hline
        \multirow{2}{*}{MirrorWiC-base} & \agsilh & \textbf{65.1} & 74.7 & 70.9 & 70.2 & 70.0 \\
        & X-Means & 63.0[±0.2] & \textbf{\textcolor{blue}{77.0}}[±1.5] & \textbf{74.4}[±0.2] & \textbf{71.2}[±0.4] & \textbf{71.6} \\
        \hline
        \hline
        \multicolumn{7}{l}{\textbf{Semi-supervised}} \\
        %\hline
        \multirow{2}{*}{BERT-l-Wikt} & \agsilh & \textbf{\textcolor{blue}{67.8}} & \textbf{75.5} & 72.4 & \textbf{\textcolor{blue}{71.8}} & \textbf{\textcolor{blue}{71.7}} \\
        & X-Means & 63.9[±1.3] & 74.9[±1.7] & \textbf{74.5}[±0.5] & 69.7[±0.3] & 71.5 \\
        %\hline
        %\multirow{2}{*}{BERT-large-Wikt-POS} & \agsilh & 67.7 & \textbf{75.5} & 72.8 & NA & \textbf{\textcolor{blue}{71.9}} \\
        %& X-Means & 65.1[±0.7] & 74.7\small[±0.4] & 73.9\small[±0.9] & NA & 71.5 \\
        \hline\hline
        \multirow{2}{*}{%\multicolumn{7}{l}{
        \textbf{Baselines}} %\\
        & 1cpl
        %\multicolumn{2}{c|}{1cpl} 
        & 65.7 & \textbf{\textcolor{red}{80.0}} & \textbf{\textcolor{red}{75.2}} & \textbf{\textcolor{red}{73.6}} & \textbf{\textcolor{red}{73.4}} \\
        & 1cpex 
        %\multicolumn{2}{c|}{1cpex} 
        & 25.5 & 22.6 & 24.1 & 24.1 & 24.2 \\
        \hline
    \end{tabular}
    \caption{{\fbcubed} performance across PLMs and clustering algorithms for each POS, for all POS (All POS), and the average over POS weighted by POS proportion in SemCor ($_w$Avg). In bold, the best value for each model supervision type (un-, self-, semi-supervised). In \textcolor{blue}{blue}, the best value for each column, excluding baselines. In \textcolor{red}{red}, cases where a baseline is best over the column. Best previous system: PolyLM-large: $73.0$ (Table \ref{tab:results_on_existing_datasets}).}
    \label{tab:semcor_baseline_results}
\end{table*}

\paragraph{Contextualized embeddings} 

We test the \textit{base-uncased} and \textit{large-uncased} versions of BERT \citep{devlin-etal-2019-BERT}\footnote{%\textit{RoBERTa-base} and \textit{RoBERTa-large} produced results equivalent or worse than their corresponding BERT models, so we omit them in the following.
For all experiments with PLMs, we use Transformers library \citep{wolf-etal-2020-transformers}. Subword embeddings are averaged per word or MWE.} ({BERT-b-u} and {BERT-l-u} hereafter). %\todo{two footnotes in a row, but talk about different things, so can't merge...}
%as well as the more recent \textit{ModernBERT-base} and \textit{ModernBERT-large} models \cite{modernBERT}
We also test two models fine-tuned for the WiC task, likely to benefit for WSI: \textit{MirrorWiC-base} \cite{liu-etal-2021-MirrorWiC}  and \textit{BERT-Wikt} \cite{mosolova-etal-2024-injecting}.
%. Both models were fine-tuned using contrastive learning, which teaches the model to bring semantically similar examples closer, while pushing dissimilar ones apart. MirrorWiC leverages self-supervised contrastive learning by using similar examples created by automatically alternating the original phrase. BERT-Wikt employs supervised contrastive learning by using exemplars of the same sense from Wiktionary as similar ones.}
%\new{ DELETE THIS:? \textit{MirrorWiC-base} \cite{liu-etal-2021-MirrorWiC}, fine-tuned using self-supervised contrastive learning, and \textit{BERT-Wikt} \cite{mosolova-etal-2024-injecting}, fine-tuned using semi-supervised contrastive learning with Wiktionary examples}
\footnote{For all our experiments, we use the dbnary dump of 06/12/2024, {\scriptsize \url{https://kaiko.getalp.org/about-dbnary/}}. As in \cite{mosolova-etal-2024-injecting}, 20\% is not used, to keep the possibility to evaluate on unused Wiktionary data.}. For the latter, we ran the fine-tuning procedure on all POS with default hyperparameters on {BERT-l-u} to obtain \textit{BERT-l-Wikt} model. For each PLM, we tested all layers and report using the best-performing layer (see Appendix \ref{app:best_layer}).% \todo{1) tell that we repeat the split of Wiktionary for adj and noun as it was done for verbs (by us in the prev. paper). this train part is used for ML constraints and data aug. as well (the dev and test are kept in case we want to evaluate on them other tasks like WiC)}

\paragraph{Clustering algorithms}

We test AG clustering with silhouette %\footnote{{\scriptsize \url{https://scikit-learn.org/stable/modules/generated/sklearn.metrics.silhouette_score.html}}} 
score (\agsilh) to determine the optimal number of clusters, and X-means, which dynamically determines the number of clusters. Being based on K-means with K++ initialization, X-means is not deterministic, we thus report the average and standard deviation of 5 runs. Hyperparameters are provided in Appendix \ref{appendix:hyper_clustering}, including default number of clusters when silhouette is not defined.

\paragraph{Handling of POS variation}

Sense distribution varies across parts of speech (cf. Table \ref{tab:semcor_statistics}). %has its own sense distribution. In WordNet \notes{you mean SemCor??}\todo{no, I took all lemmas available in WordNet and checked how many synsets they have} sense annotation, nouns have an average of 1.26[$\pm$0.92] senses, adjectives 1.41[$\pm$1.13] and verbs 2.19[$\pm$2.56]. 
To study the impact of these differences, %These differences highlight the necessity of evaluating systems separately for each POS. Thus, 
we provide results per POS and for all POS (\textbf{All POS}). Moreover, the number of lemmas for each POS is almost balanced in our subsets, but not in the full SemCor. So we also show results averaged over the 3 POS weighted by their proportions in the full SemCor (\textbf{$_w$Avg}) to reflect their natural distribution in corpus (the proportions are provided in Appendix \ref{app:pos_proportion}).%shown in Table \ref{tab:pos_proportion_semcor} in the Appendix).

\paragraph{Metrics and Statistical significance} We use {\fbcubed}, as motivated in \S \ref{sec:issues}.
% déjà dit section 3, as it is the only metric that accounts for all desirable properties outlined in \cite{Amig2008} (cf. section \ref{subsec:metrics}). Moreover, it is an item-based metric that reflects sense distribution in the corpus. 
Due to computational costs, we chose to perform the bootstrapping statistical significance test for all PLM pairs combined with {\agsilh} only and not X-means (5 reruns are needed for the latter, see details in  Appendix \ref{app:bootstrapping}).

%\todo{\textbf{significance tests}}
%%%%%%%%%%%%%%%%%%%%%%%%%%%%%%%%%%%%%%%%%%%%%%%

\subsection{Results and discussion}
\label{subsec:semcor_baselines_results}

Results are shown in Table \ref{tab:semcor_baseline_results}\footnote{ Statistical significance tests (using {\agsilh}) show that differences between all pairs of PLMs are significant at p $<0.05$, except: i) All POS: BERT-l-u versus MirrorWiC , ii) Nouns: BERT-b-u versus MirrorWiC and iii) Verbs: BERT-l-Wikt versus MirrorWiC. See also Fig. \ref{fig:bootstrapping_for_baselines_ag_all_models} in Appendix.}.

\paragraph{Baselines are high:} The first striking observation is that the 1 cluster per lemma "baseline" is actually the best technique for adjectives and nouns, and when considering all POS (All POS: $73.6$, $_{w}$Avg: $73.4$). The other systems only surpass 1cpl for verbs, namely for the most polysemous POS.

\paragraph{Models:} Among unsupervised embeddings models, {BERT-l-u} outperforms its base counterparts, overall, except for adjectives.% and \todo{\textit{ModernBERT-large}} provide the best embeddings, although the best model varies across POS. \textit{BERT-l-u} performs best for the results over all POS. 
%Regarding the clustering algorithms, in the AVG$_w$ setting, X-means performs best. 
The self-supervised finetuning of MirrorWiC surpasses the unsupervised BERTs for adjectives, but not for nouns and verbs, giving a marginal improvement overall.%the provides only a marginal improvement with respect to the BERT models. %\notes{je dirais finalement de ne pas garder MirrorWiC pour la suite?}
%Looking at results of MirrorWiC for each POS, independently of the algorithm, it performs slightly worse than unsupervised models for verbs, adjectives and nouns. However, it achieves slightly better results combining all POS (All and AVG$_w$), with X-means. It surpasses the semi-supervised BERT model for the adjectives subset. For future experiments, we will include MirrorWiC + AG$_{silh}$ results in the Appendix. 

The semi-supervised models (fine-tuned for WiC on Wiktionary) provide the best performance (excluding baselines), both for verbs and for all POS. %The benefit of using models specifically fine-tuned for each POS (BERT-large-Wikt-POS) is marginal, so we will ignore it in the following.

\paragraph{Variation across POS:} The results show that the tendencies across POS vary greatly. Using contextualized embeddings fine-tuned on Wiktionary does help in general, but not for adjectives, for which the 1cpl and then unsupervised models perform best. The tendency is opposite for verbs.

\paragraph{Clustering algorithms:} \agsilh{}  performs always better for verbs, while X-means performs always better for nouns, and most of the time for adjectives. This could be explained by X-means' tendency to define less clusters, which is beneficial for POS with lower polysemy rate. %this underestimation improves results, as the 1cpl score is relatively high for these two POS \notes{and they have lower polysemy rate compared to verbs}.
For the results over the 3 POS (All POS and $_w$Avg) X-means tends to outperform {\agsilh}, except for the semi-supervised models. Across model/algorithms, the best pair is semi-supervised model plus {\agsilh}. 

Taking these observations into account, and considering that X-means is not deterministic and needs to be run several times, %is its instability and the need to rerun each experiment multiple times. 
we will use {\agsilh} in the following experiments, the best-performing unsupervised model ({BERT-l-u}) and the semi-supervised {BERT-l-Wikt} model. %\todo{a bit weird phrase}

So for now, on a dataset more natural in terms of polysemy and sense distribution, these contextualized embeddings clustering techniques do not surpass the best previous system (PolyLM: $73.0$), and none surpasses the 1cpl technique ($73.6$). %\new{However, semi-supervised models fine-tuned for better performance on lexical semantic task outperform the original PLMs.}

%\todo{à supprimer} \notes{TODO: statistical significance: We tested statistical significance using bootstrapping (details in Appendix \ref{app:bootstrapping}) between all pairs bert-l-u and BERT-l-Wikt, for {\agsilh}, for each POS, and for "All POS". Differences result not significant for adjectives and verbs, and significant for nouns and All POS. }

%%%%%%%%%%%%%%%%%%%%%%%%%%%%%%%%%%%%%%%%%%%%%%%%%%%%%%%%%%%%%%%%%%%%%%%%%%%%%%%%%%%%%%%%%%%%%%

\section{Investigating data augmentation}
\label{sec:augmentation}

%silhouette : \url{https://scikit-learn.org/stable/modules/generated/sklearn.metrics.silhouette_score.html}

In this section, we investigate the simple technique of adding unlabeled examples to the set of instances to cluster. Augmenting the set of instances makes it denser, potentially creating new similarity links, in particular for lemmas with originally few instances. Moreover in such cases, considering more instances helps to avoid undefined silhouette cases, defaulting to one cluster (cf. Appendix \ref{appendix:hyper_clustering}). 
%Inconsistent number of examples per lemma in a more naturally distributed corpus can cause instabilities in clustering models performance. \notes{je pense pas qu'on puisse dire ça, cf. tu fais un clustering pour chaque lemme. par contre: il y a des previous works ayant utilisé le data augmentation je suppose?}To mitigate this issue, we explore data augmentation strategies aimed at enriching the dataset and improving cluster balance \notes{on ne sait pas si ça improve le cluster balance, au contraire, ça a plutôt tendance à renforcer le most frequent sense. TODO: parler plutôt de varier les contextes pour permettre à la similarité de sens de se propager de proche en proche?}. 

We investigate %two complementary approaches: 
1) unsupervised augmentation, adding either attested examples from external corpora or synthetic examples generated by LLMs; 2) semi-supervised augmentation, where we leverage Wiktionary examples for either direct dataset augmentation (based on the Wiktionary senses), and/or fine-tuning the embedding model ({BERT-l-Wikt} model \cite{mosolova-etal-2024-injecting} already used in previous sections).  

%In the following section, we detail each approach and give experimental details. 

%%%%%%%%%%%%%%%%%%%%%%%%%%%%%%%%%%%%%%%%%%%%%%%

\subsection{Dataset augmentation}

For each lemma, we augment the set of instances with either (i) Wikibooks\footnote{{\scriptsize \url{https://huggingface.co/datasets/bigscience-data/roots_en_wikibooks}}} instances, (ii) LLM-generated sentences, and (iii) Wiktionary exemplar sentences (independently of their senses).

%For each lemma, propose to augment the dataset with unlabeled examples to improve cluster balance and provide clearer decision boundaries for the clustering algorithm. These examples are sourced from 2 unsupervised and 1 semi-supervised \notes{on ne peut pas l'appeler semi-supervised , plutot lexicon??}resource. 

%\paragraph{Unsupervised sources}

For Wikibooks (WB), we extract all occurrences of all {\semcorwsi} nouns, verbs and adjectives. For each lemma, we randomly select at most N examples (N=10, 50, 100, 150). For Multiword lemmas (MWEs) examples are retrieved based on the first word of the MWE. %\notes{(to remove:)While this approach may be suboptimal for non-phrasal verbs, it provides a computationally efficient alternative to syntactic analysis for identifying lemma heads}.
We handle MWEs in the same way to retrieve the Wiktionary instances.

%While WB instances will tend to respect the sense distribution of the lemma in corpora, the Wiktionary instances will rather uniformize the sense distribution, and comprise much less instances per frequent senses in corpora. % A REMETTRE FINAL VERSION

We also test LLM-generated data.%, capitalizing on %: even though LLMs struggle to perform WSI on their own (cf. section \ref{sec:all_datasets_results}), previous work has shown 
%their lexical semantics knowledge (see Section \ref{sec:related_work}). 
%We leverage this knowledge for example generation. 
For each instance in our dataset, we provide it to the model which we prompt to generate 3 examples with same sense (the exact prompt is provided in Appendix \ref{app:prompt_generating_examples}). We use a small open-source Llama 3.1 8B 4bit and a proprietary GPT-4o\footnote{model name: gpt-4o-2024-08-06, generation seed: 42}. 
%\todo{this appendix is mentioned again in the next section, remove here or there?} \notes{I'd keep both, not the same purpose}
 Appendix \ref{app:stats_added_examples} provides the total number of added examples in each setting.%the table \ref{tab:statistics_added_examples} in the Appendix. 

%\textbf{Semi-Supervised}

%To provide semantic grounding for the clustering algorithm, we augment each lemma’s instance set with all exemplar sentences from Wiktionary. MWEs are processed identically to Wikibooks.

%we do not provide explicit semantic annotations, most Wiktionary entries include at least one example per sense. This allows the model to use them as reference points for cluster formation.

%%%%%%%%%%%%%%%%%%%%%%%%%%%%%%%%%%%%%%%%%%%%%%%

\subsection{Constrained clustering}

We extend the AG clustering algorithm by incorporating must-link constraints. For each lemma, we add all Wiktionary examples and impose must-link constraints between examples assigned to the same Wiktionary sense (assigning distance 0 for all such pairs). %\footnote{We set distances to 0 for example pairs with same Wiktionary sense. We also tested the cannot-link constraints, but not report their results as they were consistently lower than must-link. This is probably due to the higher sense granularity in Wiktionary w.r.t. Wordnet.}. A REMETTRE POUR FINAL VERSION?
For each lemma's number of clusters, we either use Wiktionnary's number of senses (\textbf{AG$_{wikt}$}), or  the silhouette score (\textbf{\agsilh}).
%, or by Wiktionary  (which might differ from the target number of clusters in {\semcorwsi}). A RAJOUTER FINAL VERSION

%%%%%%%%%%%%%%%%%%%%%%%%%%%%%%%%%%%%%%%%%%%%%%%

%\subsection{Experimental protocol}

%We experiment with different augmentation settings:

%For Wikibooks, we select 10, 50, 100 and 150 random examples per lemma.

%%%%%%%%%%%%%%%%%%%%%%%%%%%%%%%%%%%%%%%%%%%%%%%

\subsection{Results and discussion}

We conducted experiments with BERT-l-u and BERT-l-Wikt embedding models  (Tables \ref{tab:augmentation_BERT_large_uncased} and \ref{tab:augmentation_BERT_large_ft}).% and MirrorWiC-base (results in Appendix \ref{app:data_augmentation}).

%The results for \agsilh + BERT-l-u are reported in Table \ref{tab:augmentation_BERT_large_uncased}), for BERT-large-wikt-all in Table \ref{tab:augmentation_BERT_large_ft} and for MirrorWiC-base in the table \ref{tab:augmentation_MirrorWiC} in the Appendix). 

We observe the regular trend that whatever embedding model and data augmentation source, adding examples systematically improves results.

\paragraph{A new SOTA technique: } The previous SOTA system (PolyLM) requires training a masked language model from scratch. Yet, Table \ref{tab:augmentation_BERT_large_uncased} shows that it can be outperformed simply by adding sufficient unlabeled data during the clustering of contextualized embeddings: all the settings using at least 50 WikiBooks instances do surpass the PolyLM's performance ($73.0$ in Table \ref{tab:results_on_existing_datasets})\footnote{Note that PolyLM cannot benefit from data augmentation, unless by retraining a full sense embeddings model.}. 

\paragraph{Additional examples:} Comparing sources of examples, adding 10 or more WB examples per lemma results in better performance than adding Wiktionary examples, although corresponding to a similar number of examples (the exact numbers provided in Appendix \ref{app:stats_added_examples}). On the contrary, adding more than $45$k LLM-generated examples is comparable to the WB 10 examples per lemma setting. Moreover, adding more and more WB examples helps, up to the limit of 150 examples per lemma. So overall, adding attested corpus examples helps more than the litterary style examples from Wiktionary, and more than the LLM-generated examples. To conclude, adding around $100/150$ examples from raw corpora per lemma is both the cheapest and the best option. 
%Adding more unlabeled examples from WikiBooks to the clustering set consistently improves performance, up to 100 examples for the fine-tuned model and up to 150 for the unsupervised model. Augmenting the dataset with Wiktionary examples, without senses indications, also improves the results. The LLM generated examples boost the model's performance as well, with results of GPT-4o examples being slightly better. Among the unsupervised models, only those with 150 additional Wikibooks examples approach the 1cpl baseline, with Mirror-Wic + \agsilh excedding it by 0.1\% and BERT-l-u matching it (without degrading into 1cpl solution).

\paragraph{Must-link constraints:} Concerning must-link constraints, we can observe that {\agsilh} with must-link is very slightly better than without (columns 1 and 3 of Tables \ref{tab:augmentation_BERT_large_uncased} and \ref{tab:augmentation_BERT_large_ft}), except for the Wiktionary data augmentation: in the former case, we add Wiktionary examples to the other augmentation source. It seems that the improvement of must-link per se only stems from the addition of more examples.

\paragraph{Number of clusters:} Comparing columns 2 and 3 of Tables \ref{tab:augmentation_BERT_large_uncased} and \ref{tab:augmentation_BERT_large_ft}, shows that using the Wiktionary number of clusters is systematically better than silhouette (even if Wiktionary's sense inventory differs from WordNet's).
%(The most significant improvements are brought by the use of must-link constraints. This approach also benefits from addition of both generated and Wikibooks examples, but on its own, it enhances the model's performance up to 2 points (BERT-large-FT-all + AG$_{wikt}$). Generally, using the total number of senses per lemma from Wiktionary as an additional source of supervision outperforms the unsupervised \agsilh. 

\paragraph{BERT-large vs BERT-l-Wikt:} Using the fine-tuned BERT model (Table \ref{tab:augmentation_BERT_large_ft}) is always better than the corresponding BERT-large model (Table \ref{tab:augmentation_BERT_large_uncased}).

\paragraph{Comparison to the 1cpl baseline:} Finally, several settings do outperform the 1cpl baseline ($73.6$, surpassing results are shown in blue and red in Tables \ref{tab:augmentation_BERT_large_uncased} and \ref{tab:augmentation_BERT_large_ft}), \textit{\textbf{but only for settings using Wiktionary in some way}}. The best result ($75.7$) uses it in three ways (in the embedding model, the must-link constraints which also adds the Wiktionary examples, and to define the number of clusters). While this method does use a lot of manually annotated data (a full Wiktionary), we would like to emphasize that it is usable for the many languages for which a large Wiktionary exists\footnote{19 languages have Wiktionary with more than 100k entries.}. 
%\todo{to discuss: polylm, amrami; conclusion on the best models for different setting (sup, unsup)}

\begin{comment}
\begin{itemize}
    \item "tendances stables : 
    \item \gray{- must link aide}
    \item \gray{- ajouter exemples aide (wikibooks)}
    \item \gray{- nb clusters wikt mieux que nb clusters silhouette"			}
    
    \item ML/CL analysis: for some verbs cannot link doesn't let examples join despite them having one sense. the clusters depend on the word target lemma governs (truck vs truck's window, believe in God vs believe in people)		
\end{itemize}

\begin{itemize}
    \item On {\fbcubed} our system is better than SoTA systems
    \item mustlink is better than mustlink cannot link \todo{why?}
    \item {\fbcubed}: \agsilh+ML is better with BERT-large-FT-all for nouns and adjectives and for combined (for verbs, BERT-large-FT-verb is surpasses this model by 0.6\%)

    \item on {\fbcubed}, only 2 algorithms in the table combined beats 1cpv.... 
    \item on geometric mean of nmi and {\fbcubed}, we beat amrami and polylm only if use supervised settings (must link)
    \item on {\fbcubed}, we beat them on unsupervised settings as well (but all these battles are below the 1cpv baseline!)
\end{itemize}
\end{comment}

%To discuss (data augmentation):
%\begin{itemize}
%     \item  which examples are the best? 
%     \item which do not require sense annotations?
%     \item trade-off
%\end{itemize}

%\todo{do we need per pos tables?} \notes{no}

\begin{table}[h]
    \centering
    \begin{adjustbox}{max width=0.48\textwidth}
    \begin{tabular}{l|l|cc}
        \hline
        \textbf{Augmentation} & \textbf{Base} & \multicolumn{2}{c}{\textbf{Must-link}} \\
         \textbf{source} & \textbf{AG$_{silh}$} & \textbf{AG$_{wikt}$} & \textbf{AG$_{silh}$}  \\
        \hline
        None & 71.1$^\diamond$ & NA & NA  \\\hline
        Wiktionary & 72.1 & 71.9 & 72.2  \\\hline
        Llama 3.1 8B 4bit & 71.9$^\diamond$ & 73.6 & 72.7  \\
        GPT-4o & 72.4$^\diamond$ & 73.6 & 72.9  \\\hline
        WB (10 per l.) & 72.6$^\diamond$ & 73.5 & 73.0  \\
        WB (50 per l.) & 73.5$^\diamond$ & \textcolor{blue}{74.1} & \textbf{73.4}  \\
        WB (100 per l.) & 73.5$^\diamond$ & \textcolor{blue}{74.2} & 73.2 \\
        WB (150 per l.) & \textbf{73.6}$^\diamond$ & \textcolor{blue}{\textbf{74.4}} & 73.3  \\
        \hline
    \end{tabular}
    \end{adjustbox}
    \caption{{\fbcubed} results for All POS: AG clustering using BERT-l-u embeddings for various augmentation sources, with or without must-link constraints, using the nb of clusters from silhouette (\textbf{\agsilh}) or from Wiktionary (\textbf{AG$_{wikt}$}). \textbf{$^\diamond$} indicates unsupervised results, all the other ones using Wiktionary in some way. Results above 1cpl (73.6, cf. Table \ref{tab:semcor_baseline_results}) are in \textcolor{blue}{blue}.}
    \label{tab:augmentation_BERT_large_uncased}
\end{table}

\begin{table}[h]
    \centering
    \begin{adjustbox}{max width=0.48\textwidth}
    \begin{tabular}{l|l|cc}
        \hline
        \textbf{Augmentation} & \textbf{Base} & \multicolumn{2}{c}{\textbf{Must-link}} \\
         \textbf{source} & \textbf{AG$_{silh}$} & \textbf{AG$_{wikt}$} & \textbf{AG$_{silh}$} \\
        \hline
        None & 71.8 & NA & NA  \\\hline
        Wiktionary & 72.7 & \textcolor{blue}{74.1} & 72.4 \\\hline
        Llama 3.1 8B 4bit & 72.3 & \textcolor{blue}{74.7} & \textcolor{blue}{73.7} \\
        GPT-4o & 72.9 & \textcolor{blue}{75.1} & \textcolor{blue}{73.8}  \\\hline
        WB (10 per l.) & 73.6 & \textcolor{blue}{75.2} & \textcolor{blue}{73.7}  \\
        WB (50 per l.) & \textcolor{blue}{73.9} & \textcolor{blue}{75.3} & \textcolor{blue}{\textbf{73.8}} \\
        WB (100 per l.) & \textcolor{blue}{\textbf{74.5}} & \textcolor{red}{\textbf{75.7}}& \textcolor{blue}{{73.8}} \\
        WB (150 per l.) & \textcolor{blue}{74.3} & \textcolor{red}{\textbf{75.7}} & \textcolor{blue}{73.8} \\    
        \hline
    \end{tabular}
    \end{adjustbox}
    \caption{Same as Table \ref{tab:augmentation_BERT_large_uncased} but using BERT-l-Wikt. Results above 1cpl (73.6, cf. Table \ref{tab:semcor_baseline_results}) are in \textcolor{blue}{blue}.}
    \label{tab:augmentation_BERT_large_ft}
\end{table}
 
\begin{table}[H]
    \centering
    \begin{tabular}{c|c}
    \hline    
    \textbf{Model} & \textbf{\fbcubed} \\
    \hline
    PolyLM-large & 72.7 \\ 
    BERT-l-uncased+AG$_{s}$+150WB & 74.0 \\
    BERT-l-Wikt+AG$_{wikt}$+ML+100WB & \textbf{76.0} \\
    \hline
    1cpl & 74.4 \\
    \hline
        
    \end{tabular}
    \caption{{\fbcubed} results on \semcorwsi{} test set for all POS of the best previous system (cf. Table \ref{tab:results_on_existing_datasets}) and the best unsupervised and supervised models from Tables \ref{tab:augmentation_BERT_large_uncased} and \ref{tab:augmentation_BERT_large_ft} (for these models we reuse the best layer tuned on the development set).}
    \label{tab:semcor_test_results}
\end{table}

%%%%%%%%%%%%%%%%%%%%%%%%%%%%%%%%%%%%%%%%%%%%%%%%%%%%%%%%%%%%%%%%%%%%%%%%%%%%%%%%%%%%%%%%%%%%%%
\paragraph{Results on test set}
We check if the observed trends are confirmed on the test set, comparing performance for 1cpl, the best previous system (PolyLM-large), our best unsupervised system (BERT-l-u-AG$_s$+150WB) and our best Wiktionary-using system (BERT-l-Wikt-MustLink-AG$_{wikt}$+100WB). Results in Table \ref{tab:semcor_test_results} show that the Wiktionary-enhanced system ($76.0$) is best, followed by 1cpl ($74.4$), our unsupervised system ($74.0$) and previous SOTA system PolyLM ($72.7$).

\section{Conclusions}
\label{sec:concl}

In this paper, we advocated for evaluating WSI on data respecting more natural distributions of occurrences and number of senses per lemma. We proposed experiments on English, evaluated on an extract of SemCor, both reusing state-of-the-art previous methods, and investigating an LLM prompting technique, data augmentation, and a semi-supervised setting where the English Wiktionary is used in three ways (as a source for fine-tuning a BERT model on the Word-In-Context task \cite{mosolova-etal-2024-injecting}, for data augmentation and for must-link clustering constraints).

The LLM prompting technique we proposed lays far behind, calling for better leveraging the lexical semantics knowledge of LLMs. 
%However, several techniques could be explored to improve LLM performance: (i) usage of chain-of-thought prompting to first ask for a list of senses, and then to assign the instances to each of the LLM-induced senses; (ii) processing sets of instances of a given lemma in batches, followed by a merging procedure for the sense inventories induced for each batch.

%We observed systematic benefits of data augmentation on our dev and test sets, surpassing the PolyLM model \cite{ansell-etal-2021-polylm}, which was the previous SoTA system on standard WSI datasets (SE10 and SE13).

%Using Wiktionary in the three ways mentioned above does help (a result confirmed on the test set). This semi-supervised technique can be applied to the languages having a large lexicon.

Our striking conclusion is that, when considering a dataset following a more natural sense distribution and polysemy level, none of the fully unsupervised systems we tested surpass the simple baseline of clustering all instances of the same lemma into a single cluster (considering a dataset with an average polysemy close to 2, cf. Table \ref{tab:semcor_statistics}).

Simple data augmentation allows to surpass the much more complex previous SOTA model (PolyLM). BERT embeddings fine-tuned using contrastive learning on Wiktionary examples \citep{mosolova-etal-2024-injecting} are always better compared to the original BERT embeddings. More generally, we showed several ways to leverage Wiktionary allowing to surpass 1cpl on the dev and test sets (must-link constraints, data augmentation, definition of the number of clusters, fine-tuning of PLMs). Note this can be applied for the languages having a large Wiktionary or another electronic lexicon.

\begin{comment}
\item methodological issues, comparison of works difficult
\item semcorwsi, more natural
\item previous models don't beat the 1cpl, best results may vary across POS (because different sense distributions)
\item LLMs direct prompting does not work
\item data augmentation helps + wiktionary used in " ways helps, but neither previous best model (polylm) nor ours surpass on test!
\item the task is not solved
\item combination : first identify polysemous lemmas?
\item prompting with COT (lister les sense, puis demander de désambiguiser en utilisant ces sens)
\end{comment}

%\begin{comment}

\section{Future Work}
\label{sec:futurework}

We report mediocre performance, when directly prompting LLM for the WSI task. However, several techniques could be explored to improve LLM performance: (i) usage of chain-of-thought prompting to first ask for a list of senses, and then to assign the instances to each of the LLM-induced senses; (ii) processing sets of instances of a given lemma in batches, followed by a merging procedure for the sense inventories induced for each batch.

\section*{Limitations}

This paper provides an overview of the existing datasets for Word Sense Induction, introduces a new evaluation framework for this task, and uses it to establish baselines and test data augmentation techniques to improve baseline results. However, our study evaluates only three large language models and four pre-trained language models in combination with two clustering algorithms. Additionally, this research is conducted in the English language and uses only a small part of SemCor.

Despite reporting the mean and standard deviation of 5 runs for non-deterministic models, we provide the statistical significance for result differences only for some deterministic models. Since powerful statistical significance tests for {\fbcubed} involve using bootstrapped resampling, running 1000 iterations for all configurations discussed in this paper would require prohibitively expensive computational resources. 

This study was also limited by available GPU resources which included a single Nvidia A100 80GB GPU. Thus, we could not report results on the full Llama 3.3 70B model  (which requires at least 135GB of GPU memory), and instead used its quantized version.

We should have included results on adverbs as well and plan to do so for a more complete evaluation.

\section*{Acknowledgements}
We thank the reviewers for their valuable feedback on our work. 

This work has been funded by the French Agence Nationale pour la Recherche, through the SELEXINI project (ANR-21-CE23-0033-01).

\bibliography{custom}

@string{acl = {Association for Computational Linguistics}}

@string{anth = {https://aclanthology.org/}}

@string{UNIMPLICIT:2024:1 = {Proceedings of the Third Workshop on Understanding Implicit and Underspecified Language}}

@string{STARSEM:2024:1 = {Proceedings of the 13th Joint Conference on Lexical and Computational Semantics (*SEM 2024)}}

@string{LREC:2024:main = {Proceedings of the 2024 Joint International Conference on Computational Linguistics, Language Resources and Evaluation (LREC-COLING 2024)}}

@string{EMNLP:2024:main = {Proceedings of the 2024 Conference on Empirical Methods in Natural Language Processing}}

@string{EACL:2024:short = {Proceedings of the 18th Conference of the European Chapter of the Association for Computational Linguistics (Volume 2: Short Papers)}}

@string{GWC:2023:1 = {Proceedings of the 12th Global Wordnet Conference}}

@string{ACL:2022:long = {Proceedings of the 60th Annual Meeting of the Association for Computational Linguistics (Volume 1: Long Papers)}}

@string{GWC:2021:1 = {Proceedings of the 11th Global Wordnet Conference}}

@string{EACL:2021:main = {Proceedings of the 16th Conference of the European Chapter of the Association for Computational Linguistics: Main Volume}}

@string{ACL:2021:short = {Proceedings of the 59th Annual Meeting of the Association for Computational Linguistics and the 11th International Joint Conference on Natural Language Processing (Volume 2: Short Papers)}}

@string{NAACL:2019:1 = {Proceedings of the 2019 Conference of the North {A}merican Chapter of the Association for Computational Linguistics: Human Language Technologies, Volume 1 (Long and Short Papers)}}

@string{EACL:2017:1 = {Proceedings of the 15th Conference of the {E}uropean Chapter of the Association for Computational Linguistics: Volume 1, Long Papers}}

@string{LREC:2016:1 = {Proceedings of the Tenth International Conference on Language Resources and Evaluation ({LREC}`16)}}

@string{COLING:2016:1 = {Proceedings of {COLING} 2016, the 26th International Conference on Computational Linguistics: Technical Papers}}

@string{SEMEVAL:2013:2 = {Second Joint Conference on Lexical and Computational Semantics (*{SEM}), Volume 2: Proceedings of the Seventh International Workshop on Semantic Evaluation ({S}em{E}val 2013)}}

@string{IJCNLP:2013:1 = {Proceedings of the Sixth International Joint Conference on Natural Language Processing}}

@string{LREC:2012:1 = {Proceedings of the Eighth International Conference on Language Resources and Evaluation ({LREC}`12)}}

@string{SEMEVAL:2010:1 = {Proceedings of the 5th International Workshop on Semantic Evaluation}}

@string{ACL:2009:1 = {Proceedings of the Joint Conference of the 47th Annual Meeting of the {ACL} and the 4th International Joint Conference on Natural Language Processing of the {AFNLP}}}

@string{EMNLP:2009:1 = {Proceedings of the 2009 Conference on Empirical Methods in Natural Language Processing}}

@string{SEMEVAL:2007:1 = {Proceedings of the Fourth International Workshop on Semantic Evaluations ({S}em{E}val-2007)}}

@string{NAACL:2006:2 = {Proceedings of the Human Language Technology Conference of the {NAACL}, Companion Volume: Short Papers}}

@string{LREC:2004:1 = {Proceedings of the Fourth International Conference on Language Resources and Evaluation ({LREC}`04)}}

@string{HLT:1993:1 = {{H}uman {L}anguage {T}echnology: Proceedings of a Workshop Held at Plainsboro, New Jersey, March 21-24, 1993}}

@misc{Abdine_wsi,
  doi = {10.48550/ARXIV.2210.05422},
  url = {https://arxiv.org/abs/2210.05422},
  author = {Abdine,  Hadi and Eddine,  Moussa Kamal and Vazirgiannis,  Michalis and Buscaldi,  Davide},
  title = {Word Sense Induction with Hierarchical Clustering and Mutual Information Maximization},
  publisher = {arXiv},
  year = {2022},
  copyright = {arXiv.org perpetual,  non-exclusive license}
}

@inproceedings{
anonymous2025are,
title={Are Large Language Models Good Word Sense Disambiguators?},
author={Anonymous},
booktitle={Submitted to ACL Rolling Review - December 2024},
year={2025},
url={https://openreview.net/forum?id=UlzWN338K2},
note={under review}
}

@misc{ortegamartín2023linguisticambiguityanalysischatgpt,
      title={Linguistic ambiguity analysis in ChatGPT}, 
      author={Miguel Ortega-Martín and Óscar García-Sierra and Alfonso Ardoiz and Jorge Álvarez and Juan Carlos Armenteros and Adrián Alonso},
      year={2023},
      eprint={2302.06426},
      archivePrefix={arXiv},
      primaryClass={cs.CL},
      url={https://arxiv.org/abs/2302.06426}, 
}

@misc{openai2024gpt4technicalreport,
      title={GPT-4 Technical Report}, 
      author={OpenAI and Josh Achiam and Steven Adler and Sandhini Agarwal and others},
      year={2024},
      eprint={2303.08774},
      archivePrefix={arXiv},
      primaryClass={cs.CL},
      url={https://arxiv.org/abs/2303.08774}, 
}

@misc{unsloth,
  author       = {Han, Daniel and Han, Michael and Unsloth Team},
  title        = {{Unsloth}},
  year         = {2023},
  howpublished = {\url{https://github.com/unslothai/unsloth}},
  note         = {GitHub repository}
}

@misc{sumanathilaka2024llmsassistambiguityquantitative,
      title={Can LLMs assist with Ambiguity? A Quantitative Evaluation of various Large Language Models on Word Sense Disambiguation}, 
      author={T. G. D. K. Sumanathilaka and Nicholas Micallef and Julian Hough},
      year={2024},
      eprint={2411.18337},
      archivePrefix={arXiv},
      primaryClass={cs.CL},
      url={https://arxiv.org/abs/2411.18337}, 
}

@inproceedings{hayashi-2025-evaluating,
    title = "Evaluating {LLM}s' Capability to Identify Lexical Semantic Equivalence: Probing with the Word-in-Context Task",
    author = "Hayashi, Yoshihiko",
    editor = "Rambow, Owen  and
      Wanner, Leo  and
      Apidianaki, Marianna  and
      Al-Khalifa, Hend  and
      Eugenio, Barbara Di  and
      Schockaert, Steven",
    booktitle = "Proceedings of the 31st International Conference on Computational Linguistics",
    month = jan,
    year = "2025",
    address = "Abu Dhabi, UAE",
    publisher = "Association for Computational Linguistics",
    url = "https://aclanthology.org/2025.coling-main.466/",
    pages = "6985--6998"
}

@article{scikit-learn,
  title={Scikit-learn: Machine Learning in {P}ython},
  author={Pedregosa, F. and Varoquaux, G. and Gramfort, A. and Michel, V.
          and Thirion, B. and Grisel, O. and Blondel, M. and Prettenhofer, P.
          and Weiss, R. and Dubourg, V. and Vanderplas, J. and Passos, A. and
          Cournapeau, D. and Brucher, M. and Perrot, M. and Duchesnay, E.},
  journal={Journal of Machine Learning Research},
  volume={12},
  pages={2825--2830},
  year={2011},
 url={https://www.jmlr.org/papers/volume12/pedregosa11a/pedregosa11a.pdf?}
}

@article{Novikov2019,
    doi         = {10.21105/joss.01230},
    url         = {https://doi.org/10.21105/joss.01230},
    year        = 2019,
    month       = {apr},
    publisher   = {The Open Journal},
    volume      = {4},
    number      = {36},
    pages       = {1230},
    author      = {Andrei Novikov},
    title       = {PyClustering: Data Mining Library},
    journal     = {Journal of Open Source Software}
}

@book{Dror2020,
  title = {Statistical Significance Testing for Natural Language Processing},
  ISBN = {9783031021749},
  ISSN = {1947-4059},
  url = {http://dx.doi.org/10.1007/978-3-031-02174-9},
  DOI = {10.1007/978-3-031-02174-9},
  journal = {Synthesis Lectures on Human Language Technologies},
  publisher = {Springer International Publishing},
  author = {Dror,  Rotem and Peled-Cohen,  Lotem and Shlomov,  Segev and Reichart,  Roi},
  year = {2020}
}

@misc{pilehvar_wic_2019,
    title = {{WiC}: the {Word}-in-{Context} {Dataset} for {Evaluating} {Context}-{Sensitive} {Meaning} {Representations}},
    shorttitle = {{WiC}},
    url = {http://arxiv.org/abs/1808.09121},
    language = {en},
    urldate = {2023-07-27},
    publisher = {arXiv},
    author = {Pilehvar, Mohammad Taher and Camacho-Collados, Jose},
    month = apr,
    year = {2019},
    note = {arXiv:1808.09121 [cs]},
}

@article{Amplayo_Hwang_Song_2019, title={AutoSense Model for Word Sense Induction}, volume={33}, url={https://ojs.aaai.org/index.php/AAAI/article/view/4580}, DOI={10.1609/aaai.v33i01.33016212}, abstractNote={&lt;p&gt;Word sense induction (WSI), or the task of automatically discovering multiple senses or meanings of a word, has three main challenges: domain adaptability, novel sense detection, and sense granularity flexibility. While current latent variable models are known to solve the first two challenges, they are not flexible to different word sense granularities, which differ very much among words, from &lt;em&gt;aardvark&lt;/em&gt; with one sense, to &lt;em&gt;play&lt;/em&gt; with over 50 senses. Current models either require hyperparameter tuning or nonparametric induction of the number of senses, which we find both to be ineffective. Thus, we aim to eliminate these requirements and solve the &lt;strong&gt;sense granularity problem&lt;/strong&gt; by proposing &lt;strong&gt;AutoSense&lt;/strong&gt;, a latent variable model based on two observations: (1) senses are represented as a distribution over topics, and (2) senses generate pairings between the target word and its neighboring word. These observations alleviate the problem by (a) throwing garbage senses and (b) additionally inducing fine-grained word senses. Results show great improvements over the stateof-the-art models on popular WSI datasets. We also show that AutoSense is able to learn the appropriate sense granularity of a word. Finally, we apply AutoSense to the unsupervised author name disambiguation task where the sense granularity problem is more evident and show that AutoSense is evidently better than competing models. We share our data and code here: https://github.com/rktamplayo/AutoSense.&lt;/p&gt;}, number={01}, journal={Proceedings of the AAAI Conference on Artificial Intelligence}, author={Amplayo, Reinald Kim and Hwang, Seung-won and Song, Min}, year={2019}, month={Jul.}, pages={6212-6219} }

@inproceedings{sainz-etal-2023-language,title = "What do Language Models know about word senses? Zero-Shot {WSD} with Language Models and Domain Inventories",author = "Sainz, Oscar and de Lacalle, Oier Lopez and Agirre, Eneko and Rigau, German",editor = "Rigau, German and Bond, Francis and Rademaker, Alexandre",booktitle = GWC:2023:1,month = jan,year = "2023",address = "University of the Basque Country, Donostia - San Sebastian, Basque Country",publisher = "Global Wordnet Association",url = anth # {2023.gwc-1.40/},pages = "331--342"
}

@inproceedings{eyal-etal-2022-large,title = "Large Scale Substitution-based Word Sense Induction",author = "Eyal, Matan and Sadde, Shoval and Taub-Tabib, Hillel and Goldberg, Yoav",editor = "Muresan, Smaranda and Nakov, Preslav and Villavicencio, Aline",booktitle = ACL:2022:long,month = may,year = "2022",address = "Dublin, Ireland",publisher = acl,url = anth # {2022.acl-long.325/},doi = "10.18653/v1/2022.acl-long.325",pages = "4738--4752"
}

@inproceedings{manandhar-etal-2010-semeval,title = "{S}em{E}val-2010 Task 14: Word Sense Induction {\&}Disambiguation",author = "Manandhar, Suresh and Klapaftis, Ioannis and Dligach, Dmitriy and Pradhan, Sameer",editor = "Erk, Katrin and Strapparava, Carlo",booktitle = SEMEVAL:2010:1,month = jul,year = "2010",address = "Uppsala, Sweden",publisher = acl,url = anth # {S10-1011/},pages = "63--68"
}

@inproceedings{jurgens-klapaftis-2013-semeval,title = "{S}em{E}val-2013 Task 13: Word Sense Induction for Graded and Non-Graded Senses",author = "Jurgens, David and Klapaftis, Ioannis",editor = "Manandhar, Suresh and Yuret, Deniz",booktitle = SEMEVAL:2013:2,month = jun,year = "2013",address = "Atlanta, Georgia, USA",publisher = acl,url = anth # {S13-2049/},pages = "290--299"
}

@misc{amrami2019bettersubstitutionbasedwordsense,
      title={Towards better substitution-based word sense induction}, 
      author={Asaf Amrami and Yoav Goldberg},
      year={2019},
      eprint={1905.12598},
      archivePrefix={arXiv},
      primaryClass={cs.CL},
      url={https://arxiv.org/abs/1905.12598}, 
}

@inproceedings{blevins-etal-2021-fews,title = "{FEWS}: Large-Scale, Low-Shot Word Sense Disambiguation with the Dictionary",author = "Blevins, Terra and Joshi, Mandar and Zettlemoyer, Luke",editor = "Merlo, Paola and Tiedemann, Jorg and Tsarfaty, Reut",booktitle = EACL:2021:main,month = apr,year = "2021",address = "Online",publisher = acl,url = anth # {2021.eacl-main.36/},doi = "10.18653/v1/2021.eacl-main.36",pages = "455--465"
}

@inproceedings{devlin-etal-2019-bert,title = "{BERT}: Pre-training of Deep Bidirectional Transformers for Language Understanding",author = "Devlin, Jacob and Chang, Ming-Wei and Lee, Kenton and Toutanova, Kristina",editor = "Burstein, Jill and Doran, Christy and Solorio, Thamar",booktitle = NAACL:2019:1,month = jun,year = "2019",address = "Minneapolis, Minnesota",publisher = acl,url = anth # {N19-1423/},doi = "10.18653/v1/N19-1423",pages = "4171--4186"
}

@inproceedings{lietard-etal-2024-word,title = "To Word Senses and Beyond: Inducing Concepts with Contextualized Language Models",author = "Li{\'e}tard, Bastien and Denis, Pascal and Keller, Mikaela",editor = "Al-Onaizan, Yaser and Bansal, Mohit and Chen, Yun-Nung",booktitle = EMNLP:2024:main,month = nov,year = "2024",address = "Miami, Florida, USA",publisher = acl,url = anth # {2024.emnlp-main.156/},doi = "10.18653/v1/2024.emnlp-main.156",pages = "2684--2696"
}

@inproceedings{yava-etal-2024-improving,title = "Improving Word Sense Induction through Adversarial Forgetting of Morphosyntactic Information",author = "Yavas, Deniz Ekin and Bernard, Timoth{\'e}e and Kallmeyer, Laura and Crabb{\'e}, Beno{\^i}t",editor = "Bollegala, Danushka and Shwartz, Vered",booktitle = STARSEM:2024:1,month = jun,year = "2024",address = "Mexico City, Mexico",publisher = acl,url = anth # {2024.starsem-1.19/},doi = "10.18653/v1/2024.starsem-1.19",pages = "238--251"
}

@inproceedings{ansell-etal-2021-polylm,
    title = "{P}oly{LM}: Learning about Polysemy through Language Modeling",
    author = "Ansell, Alan  and
      Bravo-Marquez, Felipe  and
      Pfahringer, Bernhard",
    editor = "Merlo, Paola  and
      Tiedemann, Jorg  and
      Tsarfaty, Reut",
    booktitle = "Proceedings of the 16th Conference of the European Chapter of the Association for Computational Linguistics: Main Volume",
    month = apr,
    year = "2021",
    address = "Online",
    publisher = "Association for Computational Linguistics",
    url = "https://aclanthology.org/2021.eacl-main.45/",
    doi = "10.18653/v1/2021.eacl-main.45",
    pages = "563--574"
}

@inproceedings{periti-etal-2024-automatically,title = "Automatically Generated Definitions and their utility for Modeling Word Meaning",author = "Periti, Francesco and Alfter, David and Tahmasebi, Nina",editor = "Al-Onaizan, Yaser and Bansal, Mohit and Chen, Yun-Nung",booktitle = EMNLP:2024:main,month = nov,year = "2024",address = "Miami, Florida, USA",publisher = acl,url = anth # {2024.emnlp-main.776/},doi = "10.18653/v1/2024.emnlp-main.776",pages = "14008--14026"
}

@article{Amig2008,
  title = {A comparison of extrinsic clustering evaluation metrics based on formal constraints},
  volume = {12},
  ISSN = {1573-7659},
  url = {http://dx.doi.org/10.1007/s10791-008-9066-8},
  DOI = {10.1007/s10791-008-9066-8},
  number = {4},
  journal = {Information Retrieval},
  publisher = {Springer Science and Business Media LLC},
  author = {Amigó,  Enrique and Gonzalo,  Julio and Artiles,  Javier and Verdejo,  Felisa},
  year = {2008},
  month = jul,
  pages = {461–486}
}

@inproceedings{biemann-2012-turk,title = "Turk Bootstrap Word Sense Inventory 2.0: A Large-Scale Resource for Lexical Substitution",author = "Biemann, Chris",editor = "Calzolari, Nicoletta and Choukri, Khalid and Declerck, Thierry and Do{\u{g}}an, Mehmet U{\u{g}}ur and Maegaard, Bente and Mariani, Joseph and Moreno, Asuncion and Odijk, Jan and Piperidis, Stelios",booktitle = LREC:2012:1,month = may,year = "2012",address = "Istanbul, Turkey",publisher = "European Language Resources Association (ELRA)",url = anth # {L12-1101/},pages = "4038--4042"
}

@inproceedings{erk-etal-2009-investigations,title = "Investigations on Word Senses and Word Usages",author = "Erk, Katrin and McCarthy, Diana and Gaylord, Nicholas",editor = "Su, Keh-Yih and Su, Jian and Wiebe, Janyce and Li, Haizhou",booktitle = ACL:2009:1,month = aug,year = "2009",address = "Suntec, Singapore",publisher = acl,url = anth # {P09-1002/},pages = "10--18"
}

@inproceedings{erk-mccarthy-2009-graded,title = "Graded Word Sense Assignment",author = "Erk, Katrin and McCarthy, Diana",editor = "Koehn, Philipp and Mihalcea, Rada",booktitle = EMNLP:2009:1,month = aug,year = "2009",address = "Singapore",publisher = acl,url = anth # {D09-1046/},pages = "440--449"
}

@inproceedings{habibi-etal-2021-homonymy,title = "Homonymy and Polysemy Detection with Multilingual Information",author = "Habibi, Amir Ahmad and Hauer, Bradley and Kondrak, Grzegorz",editor = "Vossen, Piek and Fellbaum, Christiane",booktitle = GWC:2021:1,month = jan,year = "2021",address = "University of South Africa (UNISA)",publisher = "Global Wordnet Association",url = anth # {2021.gwc-1.4/},pages = "26--35"
}

@inproceedings{herman-jakubicek-2024-shadowsense,title = "{S}hadow{S}ense: A Multi-annotated Dataset for Evaluating Word Sense Induction",author = "Herman, Ond{\v{r}}ej and Jakub{\'i}{\v{c}}ek, Milo{\v{s}}",editor = "Calzolari, Nicoletta and Kan, Min-Yen and Hoste, Veronique and Lenci, Alessandro and Sakti, Sakriani and Xue, Nianwen",booktitle = LREC:2024:main,month = may,year = "2024",address = "Torino, Italia",publisher = "ELRA and ICCL",url = anth # {2024.lrec-main.1286/},pages = "14763--14769"
}

@inproceedings{hovy-etal-2006-ontonotes,title = "{O}nto{N}otes: The 90{\%} Solution",author = "Hovy, Eduard and Marcus, Mitchell and Palmer, Martha and Ramshaw, Lance and Weischedel, Ralph",editor = "Moore, Robert C. and Bilmes, Jeff and Chu-Carroll, Jennifer and Sanderson, Mark",booktitle = NAACL:2006:2,month = jun,year = "2006",address = "New York City, USA",publisher = acl,url = anth # {N06-2015/},pages = "57--60"
}

@inproceedings{ide-suderman-2004-american,title = "The {A}merican National Corpus First Release",author = "Ide, Nancy and Suderman, Keith",editor = "Lino, Maria Teresa and Xavier, Maria Francisca and Ferreira, F{\'a}tima and Costa, Rute and Silva, Raquel",booktitle = LREC:2004:1,month = may,year = "2004",address = "Lisbon, Portugal",publisher = "European Language Resources Association (ELRA)",url = anth # {L04-1313/}
}

@article{li-etal-2014-improved,title = "Improved Estimation of Entropy for Evaluation of Word Sense Induction",author = "Li, Linlin and Titov, Ivan and Sporleder, Caroline",journal = "Computational Linguistics",volume = "40",number = "3",month = sep,year = "2014",address = "Cambridge, MA",publisher = "MIT Press",url = anth # {J14-3007/},doi = "10.1162/COLI_a_00196",pages = "671--685"
}

@inproceedings{yavas-2024-assessing,title = "Assessing the Significance of Encoded Information in Contextualized Representations to Word Sense Disambiguation",author = "Yavas, Deniz Ekin",editor = "Pyatkin, Valentina and Fried, Daniel and Stengel-Eskin, Elias and Liu, Alisa and Pezzelle, Sandro",booktitle = UNIMPLICIT:2024:1,month = mar,year = "2024",address = "Malta",publisher = acl,url = anth # {2024.unimplicit-1.4/},pages = "42--53"
}

@inproceedings{springorum-etal-2013-detecting,title = "Detecting Polysemy in Hard and Soft Cluster Analyses of {G}erman Preposition Vector Spaces",author = "Springorum, Sylvia and Schulte im Walde, Sabine and Utt, Jason",editor = "Mitkov, Ruslan and Park, Jong C.",booktitle = IJCNLP:2013:1,month = oct,year = "2013",address = "Nagoya, Japan",publisher = "Asian Federation of Natural Language Processing",url = anth # {I13-1072/},pages = "632--640"
}

@inproceedings{lossio-ventura-etal-2016-automatic,title = "Automatic Biomedical Term Polysemy Detection",author = "Lossio-Ventura, Juan Antonio and Jonquet, Clement and Roche, Mathieu and Teisseire, Maguelonne",editor = "Calzolari, Nicoletta and Choukri, Khalid and Declerck, Thierry and Goggi, Sara and Grobelnik, Marko and Maegaard, Bente and Mariani, Joseph and Mazo, Helene and Moreno, Asuncion and Odijk, Jan and Piperidis, Stelios",booktitle = LREC:2016:1,month = may,year = "2016",address = "Portoro{\v{z}}, Slovenia",publisher = "European Language Resources Association (ELRA)",url = anth # {L16-1266/},pages = "1684--1688"
}

@inproceedings{miller-etal-1993-semantic,title = "A Semantic Concordance",author = "Miller, George A. and Leacock, Claudia and Tengi, Randee and Bunker, Ross T.",booktitle = HLT:1993:1,year = "1993",url = anth # {H93-1061/}
}

@inproceedings{mihalcea-etal-2004-senseval,
    title = "The Senseval-3 {E}nglish lexical sample task",
    author = "Mihalcea, Rada  and
      Chklovski, Timothy  and
      Kilgarriff, Adam",
    booktitle = "Proceedings of {SENSEVAL}-3, the Third International Workshop on the Evaluation of Systems for the Semantic Analysis of Text",
    month = jul,
    year = "2004",
    address = "Barcelona, Spain",
    publisher = "Association for Computational Linguistics",
    url = "https://aclanthology.org/W04-0807/",
    pages = "25--28"
}

@inproceedings{wolf-etal-2020-transformers,
    title = "Transformers: State-of-the-Art Natural Language Processing",
    author = "Thomas Wolf and Lysandre Debut and Victor Sanh and Julien Chaumond and Clement Delangue and Anthony Moi and Pierric Cistac and Tim Rault and Rémi Louf and Morgan Funtowicz and Joe Davison and Sam Shleifer and Patrick von Platen and Clara Ma and Yacine Jernite and Julien Plu and Canwen Xu and Teven Le Scao and Sylvain Gugger and Mariama Drame and Quentin Lhoest and Alexander M. Rush",
    booktitle = "Proceedings of the 2020 Conference on Empirical Methods in Natural Language Processing: System Demonstrations",
    month = oct,
    year = "2020",
    address = "Online",
    publisher = "Association for Computational Linguistics",
    url = "https://www.aclweb.org/anthology/2020.emnlp-demos.6",
    pages = "38--45"
}

@inproceedings{liu-etal-2021-mirrorwic,
    title = "{M}irror{W}i{C}: On Eliciting Word-in-Context Representations from Pretrained Language Models",
    author = "Liu, Qianchu  and
      Liu, Fangyu  and
      Collier, Nigel  and
      Korhonen, Anna  and
      Vuli{\'c}, Ivan",
    editor = "Bisazza, Arianna  and
      Abend, Omri",
    booktitle = "Proceedings of the 25th Conference on Computational Natural Language Learning",
    month = nov,
    year = "2021",
    address = "Online",
    publisher = "Association for Computational Linguistics",
    url = "https://aclanthology.org/2021.conll-1.44/",
    doi = "10.18653/v1/2021.conll-1.44",
    pages = "562--574"
}

@misc{grattafiori2024llama3herdmodels,
      title={The Llama 3 Herd of Models}, 
      author={Aaron Grattafiori and Abhimanyu Dubey and Abhinav Jauhri and Abhinav Pandey and others},
      year={2024},
      eprint={2407.21783},
      archivePrefix={arXiv},
      primaryClass={cs.AI},
      url={https://arxiv.org/abs/2407.21783}, 
      note = "arXiv~: \href{https://arxiv.org/abs/2407.21783}{2407.21783}"
}

@inproceedings{schlechtweg-etal-2020-semeval,
    title = "{S}em{E}val-2020 Task 1: Unsupervised Lexical Semantic Change Detection",
    author = "Schlechtweg, Dominik  and
      McGillivray, Barbara  and
      Hengchen, Simon  and
      Dubossarsky, Haim  and
      Tahmasebi, Nina",
    editor = "Herbelot, Aurelie  and
      Zhu, Xiaodan  and
      Palmer, Alexis  and
      Schneider, Nathan  and
      May, Jonathan  and
      Shutova, Ekaterina",
    booktitle = "Proceedings of the Fourteenth Workshop on Semantic Evaluation",
    month = dec,
    year = "2020",
    address = "Barcelona (online)",
    publisher = "International Committee for Computational Linguistics",
    url = "https://aclanthology.org/2020.semeval-1.1/",
    doi = "10.18653/v1/2020.semeval-1.1",
    pages = "1--23"
}

@inproceedings{panchenko-etal-2017-unsupervised-mean,title = "Unsupervised Does Not Mean Uninterpretable: The Case for Word Sense Induction and Disambiguation",author = "Panchenko, Alexander and Ruppert, Eugen and Faralli, Stefano and Ponzetto, Simone Paolo and Biemann, Chris",editor = "Lapata, Mirella and Blunsom, Phil and Koller, Alexander",booktitle = EACL:2017:1,month = apr,year = "2017",address = "Valencia, Spain",publisher = acl,url = anth # {E17-1009/},pages = "86--98"
}

@inproceedings{yamada-etal-2021-semantic,title = "Semantic Frame Induction using Masked Word Embeddings and Two-Step Clustering",author = "Yamada, Kosuke and Sasano, Ryohei and Takeda, Koichi",editor = "Zong, Chengqing and Xia, Fei and Li, Wenjie and Navigli, Roberto",booktitle = ACL:2021:short,month = aug,year = "2021",address = "Online",publisher = acl,url = anth # {2021.acl-short.102/},doi = "10.18653/v1/2021.acl-short.102",pages = "811--816"
}

@inproceedings{xypolopoulos-etal-2021-unsupervised,title = "Unsupervised Word Polysemy Quantification with Multiresolution Grids of Contextual Embeddings",author = "Xypolopoulos, Christos and Tixier, Antoine and Vazirgiannis, Michalis",editor = "Merlo, Paola and Tiedemann, Jorg and Tsarfaty, Reut",booktitle = EACL:2021:main,month = apr,year = "2021",address = "Online",publisher = acl,url = anth # {2021.eacl-main.297/},doi = "10.18653/v1/2021.eacl-main.297",pages = "3391--3401"
}

@inproceedings{mosolova-etal-2024-injecting,title = "Injecting {W}iktionary to improve token-level contextual representations using contrastive learning",author = "Mosolova, Anna and Candito, Marie and Ramisch, Carlos",editor = "Graham, Yvette and Purver, Matthew",booktitle = EACL:2024:short,month = mar,year = "2024",address = "St. Julian{'}s, Malta",publisher = acl,url = anth # {2024.eacl-short.5/},pages = "34--41"
}

@inproceedings{komninos-manandhar-2016-structured,title = "Structured Generative Models of Continuous Features for Word Sense Induction",author = "Komninos, Alexandros and Manandhar, Suresh",editor = "Matsumoto, Yuji and Prasad, Rashmi",booktitle = COLING:2016:1,month = dec,year = "2016",address = "Osaka, Japan",publisher = "The COLING 2016 Organizing Committee",url = anth # {C16-1337/},pages = "3577--3587"
}

@inproceedings{agirre-soroa-2007-semeval,title = "{S}em{E}val-2007 Task 02: Evaluating Word Sense Induction and Discrimination Systems",author = "Agirre, Eneko and Soroa, Aitor",editor = "Agirre, Eneko and M{\`a}rquez, Llu{\'i}s and Wicentowski, Richard",booktitle = SEMEVAL:2007:1,month = jun,year = "2007",address = "Prague, Czech Republic",publisher = acl,url = anth # {S07-1002/},pages = "7--12"
}

@inproceedings{navigli-vannella-2013-semeval,title = "{S}em{E}val-2013 Task 11: Word Sense Induction and Disambiguation within an End-User Application",author = "Navigli, Roberto and Vannella, Daniele",editor = "Manandhar, Suresh and Yuret, Deniz",booktitle = SEMEVAL:2013:2,month = jun,year = "2013",address = "Atlanta, Georgia, USA",publisher = acl,url = anth # {S13-2035/},pages = "193--201"
}
%\bibliography{custom}

%%%%%%%%%%%%%%%%%%%%%%%%%%%%%%%%%%%%%%%%%%%%%%%%%%%%%%%%%%%%%%%%%%%%%%%%%%%%%%%%%%%%

\appendix

\section{WSI datasets}
\label{app:wsidatasets}
%\notes{ne garder que les détails sur lemma and instance selection, and sense}
%%%%%%%%%%%%%%%%%%%%%%%%%%%%%%%%%%%%%%%%%%

\subsection{SemEval 2010 Task 14}
\citet{manandhar-etal-2010-SemEval} propose a WSI task, the goal of which is to train a system on an unannotated corpus and then use it to annotate test instances of the lemmas present in the training corpus. The dataset contains instances of 100 lemmas. The unsupervised training set is composed of contexts for each lemma obtained through automated queries using WordNet-related word lemmas. The test set contains unseen instances of each lemma originating from OntoNotes \citep{hovy-etal-2006-ontonotes} annotated with OntoNotes senses. We note that, at the time, OntoNotes had an annotation only of the most frequent polysemous lemmas within a subset of PropBank. Statistics on the dataset size are given in Table \ref{tab:SemEval2010_dataset}. 

.%\footnote{\todo{delete footnote after}: 1) there are some senses of some lemmas that have only one example in the test set, but each lemma has at least 2 senses , 2) the dataset is a part of the first version of Ontonotes, citation from Ontonotes paper: \textit{"As part of OntoNotes we are annotating the most frequent noun and verb senses in a 300K subset of the PropBank, and will have this data available for release in early 2007."}$\rightarrow$ examples do not come from entire texts, but from small pieces found for the most frequent lemmas} 

\begin{table}[!h]
  \centering
  \begin{adjustbox}{max width=0.5\textwidth}

  \begin{tabular}{cccc}
    \hline
    & \textbf{Training set} & \textbf{Testing set} & \textbf{Senses (AVG)} \\
    \hline
        All & 879807 & 8915 & 3.79 \\
        Nouns & 716945 & 5285 & 4.46 \\
        Verbs & 162862 & 3630 & 3.12 \\
    \hline
  \end{tabular}
  \end{adjustbox}
  \caption{Training \& testing set details from SemEval 2010 paper \cite{manandhar-etal-2010-SemEval}.}
  \label{tab:SemEval2010_dataset}
\end{table}

%\notes{on peut enlever tout ça: This task uses two types of evaluation: supervised (with remapping of the induced senses) and unsupervised. In this paper, we consider only unsupervised metrics, namely: V-measure and Paired F-score. V-measure \cite{rosenberg-hirschberg-2007-v} is an entropy-based metric defined as the harmonic mean of homogeneity and completeness. Homogeneity measures the degree that each cluster consists of data points primarily belonging to a single class, \todo{while completeness evaluates the inverse (poor wording)}. Paired F-score \citep{artiles-etal-2009-role} is a pair-counting-based metric defined as the harmonic mean of precision and recall, where precision is the ratio of correct pairs in the clustering solution to the total number of pairs in the clustering solution and recall is the number of correct pairs divided by the total number of pairs in the gold labels \todo{too many words about the same}. The total of both metrics is calculated as the weighted mean of the result for each lemma to account for imbalance in the number of instances for each lemma.}

%%%%%%%%%%%%%%%%%%%%%%%%%%%%%%%%%%%%%%%%%%

\subsection{SemEval 2013 Task 13}
\citet{jurgens-klapaftis-2013-SemEval} introduced a new task which consists in annotating instances of lemmas with one or more senses and weighting each by their applicability (Graded Word Sense Induction). The dataset is divided into two parts: a trial set and a testing set. The trial set includes 8 lemmas, each with 50 contexts (data gathered by \citet{erk-mccarthy-2009-graded}). The test set contains 50 lemmas, with each lemma having between 22 and 100 contexts, annotated using WordNet senses. An important consideration for this dataset is the nature and the annotation difference for the trial and testing sets. The trial set is composed of a mix of 25 SemCor \citep{miller-etal-1993-semantic} and 25 SENSEVAL-3 \citep{mihalcea-etal-2004-senseval} random examples, while the test set was gathered from the Open American National Corpus \citep{ide-suderman-2004-american}. The annotation process of the trial set was performed by three untrained lexicographers who evaluated the applicability of each WordNet sense on a 5 point scale \cite{erk-etal-2009-investigations}, while the testing set was annotated by the authors of the paper \cite{jurgens-klapaftis-2013-SemEval} on a 4 point scale. %The difference in the annotation process resulted in high difference between the average number of senses per instance between the trial and testing sets resulting in decreased scores of some participants of the initial SemEval Task. 
The dataset statistics are presented in Table \ref{tab:SemEval2013_dataset}. 

%\notes{maybe add: Additional unsupervised corpus: the ukWaC corpus (Baroni et al., 2009) }

\begin{table}[!h]
  \centering
  \begin{tabular}{ccc}
    \hline
     & \textbf{Testing set} & \textbf{Trial set} \\
    \hline
        Instances & 4664 & 400 \\
        AVG senses/inst. & 1.12 & 4.97 \\
    \hline
  \end{tabular}
  \caption{Testing set details from SemEval 2013 paper \cite{jurgens-klapaftis-2013-SemEval}, trial set details computed on the provided dataset. \textbf{AVG senses/inst.}: mean number of applicable senses per instance.}
  \label{tab:SemEval2013_dataset}
\end{table}

\subsection{Other WSI Datasets}
\label{app:subsec:other}

Other datasets for WSI evaluation include SemEval 2007 Task 2 \citep{agirre-soroa-2007-SemEval}, which is replaced by SemEval 2010 (and has similar issues), SemEval 2013 Task 11 \citep{navigli-vannella-2013-SemEval} on clustering web query results, and corresponding to WSI when queries contain single words, and the aforementioned CoNLL-2025 Robust WSI Task, whose final evaluation data was not yet released.

Beyond shared tasks, some authors of WSI models proposed their own datasets and metrics because of the issues discussed in Section~\ref{sec:issues}. 
\citet{eyal-etal-2022-large} create their own dataset by annotating 20 ambiguous lemmas each represented by 100 random contexts from English Wikipedia to evaluate their large scale system across sense-induced Wikipedia. This dataset is not available online, therefore we did not report results on it. \citet{panchenko-etal-2017-unsupervised-mean} used SemEval 2013 Task 13 and the dataset proposed by \citet{biemann-2012-turk} to evaluate different configurations of their system. This dataset later was not used by other authors, it is also out of scope for our paper.

%\car{I did some rearrangement of what stays in section 3 and what moves to appendix + reformulated Other WSI Datasets check if OK}

%%%%%%%%%%%%%%%%%%%%%%%%%%%%%%%%%%%%%%%%%%%%%%%%%%%%%%%%%%%

\begin{table*}[!h]
    \centering
    \begin{adjustbox}{max width=\textwidth}
    \begin{tabular}{c|ccc|ccc|ccc|ccc|cc}
    \textbf{Metric} & \multicolumn{3}{c|}{\textbf{Homogeneity}} & \multicolumn{3}{c|}{\textbf{Completeness}} & \multicolumn{3}{c|}{\textbf{Rag Bag}} & \multicolumn{3}{c|}{\textbf{Size vs Quality}} & \textbf{1cpl} & \textbf{1cpex} \\

    & \multicolumn{3}{c|}{\includegraphics[width=0.2\linewidth]{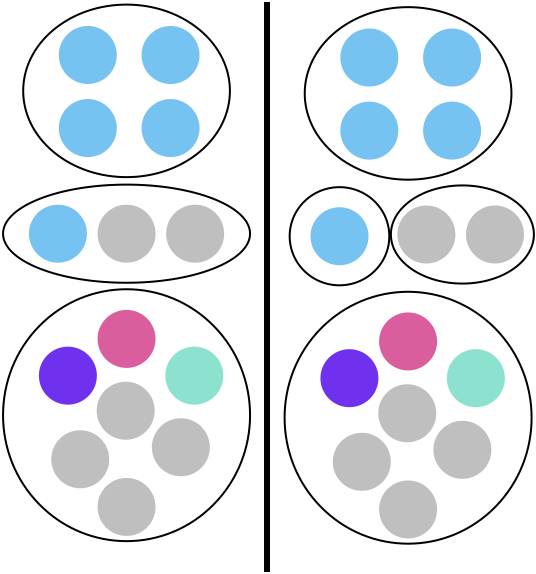}} 
    & \multicolumn{3}{c|}{\includegraphics[width=0.2\linewidth]{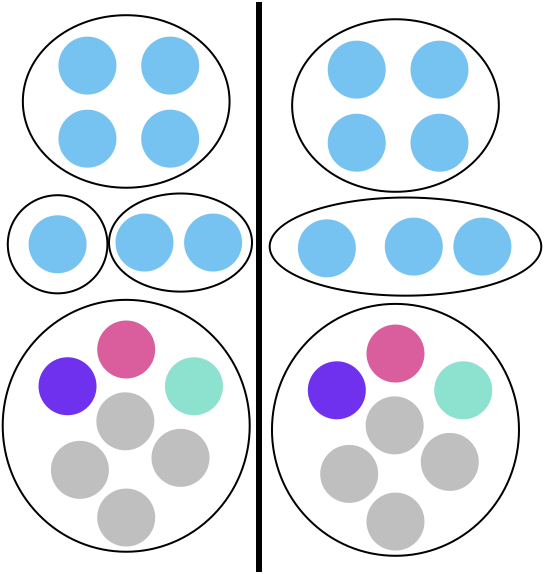}} 
    & \multicolumn{3}{c|}{\includegraphics[width=0.2\linewidth]{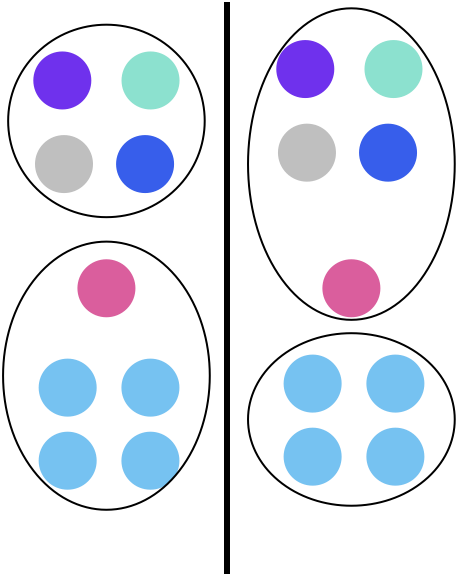}} 
    & \multicolumn{3}{c|}{\includegraphics[width=0.2\linewidth]{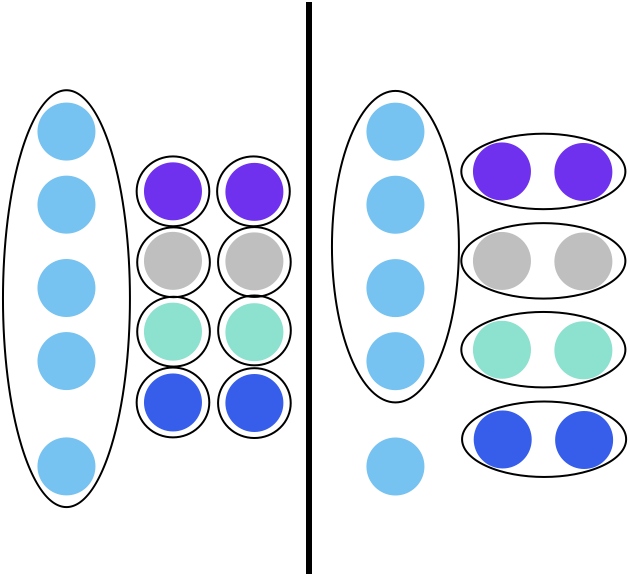}} 
    & \includegraphics[width=0.1\linewidth]{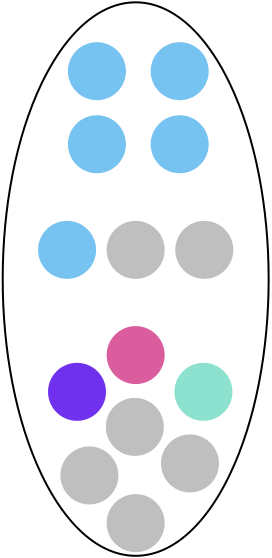} 
    & \includegraphics[width=0.1\linewidth]{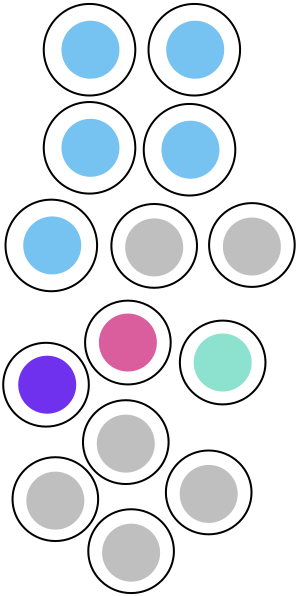} \\
    
    %& v1 & v2 & diff &  v1 & v2 & diff & v1 & v2 & diff & v1 & v2 & diff & & \\
    \hline
    Rand index & 0.68 & 0.7 & $\surd$ & 0.68 & 0.7 & $\surd$ & 0.72 & 0.72 & $\times$ & 0.95 & 0.95 & $\times$ & 0.27 & 0.73 \\
    Paired F-score & 0.47 & 0.49 & $\surd$ & 0.47 & 0.53 & $\surd$ & 0.55 & 0.55 & $\times$ & 0.83 & 0.83 & $\times$ & 0.43 & ND \\
    \hline
    NMI & 0.45 & 0.56 & $\surd$ & 0.55 & 0.55 & $\times$ & 0.43 & 0.43 & $\times$ & 0.78 & 0.89 & $\surd$ & 0.0 & 0.49 \\
    V-measure & 0.5 & 0.58 & $\surd$ & 0.57 & 0.6 & $\surd$ & 0.61 & 0.61 & $\times$ & 0.88 & 0.94 & $\surd$ & 0.0 & 0.66 \\
    \hline
    B$^3$ Precision & 0.6 & 0.69 & $\surd$ & 0.69 & 0.69 & $\times$ & 0.49 & 0.56 & $\surd$ & 1.0 & 1.0 & $\times$ & 0.33 & 1.0 \\
    B$^3$ Recall & 0.7 & 0.7 & $\times$ & 0.71 & 0.76 & $\surd$ & 1.0 & 1.0 & $\times$ & 0.69 & 0.88 & $\surd$ & 1.0 & 0.36 \\
    F-B$^3$ & 0.64 & 0.69 & $\surd$ & 0.7 & 0.72 & $\surd$ & 0.66 & 0.71 & $\surd$ & 0.82 & 0.93 & $\surd$ & 0.49 & 0.53 \\
    
    \end{tabular}
    \end{adjustbox}
    \caption{Verification of clustering metrics properties proposed by \citet{Amig2008} on their use cases (see figure 11 in their paper) for the metrics discussed in the \S \ref{sec:issues}. We also test the metrics on the 1 cluster per lemma and 1 cluster per instance solutions.}
    \label{tab:amigo_new_metrics}
\end{table*}

\section{Evaluation metrics properties}
\label{app:metrics-properties}

Table~\ref{tab:amigo_new_metrics} presents our replication of the analysis by \citet{Amig2008}, including WSI simple baselines: one cluster per lemma and one clsuer per instance.

%%%%%%%%%%%%%%%%%%%%%%%%%%%%%%%%%%%%%%%%%%%%%%%%%%%%%%%%%%%

\section{GPU usage}
\label{app:gpu_usage}

For all experiments described in this paper, we used a single Nvidia A100 80GB GPU card. In Table \ref{app:gpu_usage}, we detail the GPU memory requirements and processing times for each model when using the {\semcorwsi} dataset. The total GPU computational time for all experiments reported in this paper is 67.5 hours, excluding the time spent on statistical significance tests. The experiments described in \S\ref{sec:variability} required 2 hours for both PolyLM and LSDP models, 34 hours for direct prompting of Llama models, and 7.5 hours for GPT-4o. The experiments in \S\ref{sec:techniques} took 4 hours, while those in \S\ref{sec:augmentation} required 20 hours.

\begin{table}[H]
    \centering
    \begin{tabular}{ccc}
    \hline
    \textbf{Model} &\textbf{Size on GPU} & $\mathbf{\tau}$ \\
    \hline
       Llama 3.1 8B & 16GB & 40m \\
       Llama 3.3 70B & 135GB & - \\
       Llama 3.3 70B 4bit & 38GB & 3h40m \\
       BERT-large\tablefootnote{Number of parameters of BERT-large-uncased: 334M} + AG$_{s}$ & 3GB & 5m45s \\ 
       GPT-4o &  UNK & 45 min\\
    \hline
    \end{tabular}
    \caption{GPU usage of each model, where \textbf{$\tau$} represents the {\semcorwsi} processing time. \textbf{Size on GPU} indicates the model's size, which may double during inference. All values are approximate and may vary slightly.}
    \label{tab:gpu_usage}
\end{table}

\section{Modification of the \citet{amrami2019bettersubstitutionbasedwordsense} algorithm}
\label{app:amrami}

We modified the process of determining strong and weak senses of the LSDP algorithm. Specifically, strong senses are defined as senses that dominate at least 2 instances (with 2 a hyperparameter tuned by the authors). When a lemma has less instances than the default number of senses, a scenario not occurring for the data tested by the authors, it is possible that each sense would be dominant exactly once (or less). We tested two strategies to manage this scenario: 1) consider all senses as weak, clustering all instances together, and 2) consider all senses as strong, placing them in separate clusters. The former strategy yielded better results, and thus, we report only them in the table \ref{tab:semcor_baseline_results}. 

%%%%%%%%%%%%%%%%%%%%%%%%%%%%%%%%%%%%%%%%%%%%%%%%%%%%%%%%%%%

\section{\semcorwsi{} dataset statistics}
\label{app:semcorwsi_statistics}

In Table \ref{tab:semcor_statistics}, the statistics of the \semcorwsi{} dataset are provided in comparison with the original SemCor corpus.

\begin{table}[!ht]
    \centering
    \begin{adjustbox}{max width=0.48\textwidth}
    \begin{tabular}{@{}c@{~~}c@{~~}c@{~~}c|c@{~~}c@{~~}c|c@{}}
        \hline
        POS & \multicolumn{3}{c|}{\textbf{Dev set}} & \multicolumn{3}{c|}{\textbf{Test set}} & \textbf{SemCor} \\
        & Inst. & Lem. & Polysemy & Inst. & Lem. & Polysemy & Polysemy \\
        \hline
        %Adj & 5298 & 822 & 1.36[$\pm$1.03] & 5168 & 823 & 1.36[$\pm$0.97] \\
        %Nouns  & 5780 & 865 & 1.42[$\pm$1.12] & 6066 & 865 & 1.42[$\pm$1.13] \\
        %Verbs & 5218 & 572 & 1.87[$\pm$1.86] & 5184 & 572 & 1.87[$\pm$2.0] \\
        %All & 16296 & 2259 & 1.53[$\pm$1.38] & 16418 & 2260 & 1.52[$\pm$1.4] \\ 
        %\multicolumn{7}{c}{w/o singles} \\
        Adj & 4909 & 433 & 1.69\small[$\pm$1.3] &  4772 & 427 & 1.69\small[$\pm$1.2] & 1.64\small[$\pm$1.2] \\ 
        Noun & 5394 & 479 & 1.75\small[$\pm$1.4] & 5694 & 493 & 1.73\small[$\pm$1.4]  & 1.71\small[$\pm$1.4] \\
        Verb & 5005 & 359 & 2.39\small[$\pm$2.2] & 4979 & 367 & 2.36\small[$\pm$2.4] & 2.34\small[$\pm$2.5] \\
        All & 15308 & 1271 & 1.94\small[$\pm$1.7] & 15445 & 1287 & 1.91\small[$\pm$1.7] & 2.1\small[$\pm$2.2] \\
        
        \hline
    \end{tabular}
    \end{adjustbox}
    \caption{{\semcorwsi} dataset statistics (no hapaxes).} %\todo{@Marie, if smaller std values in the last column are ok, I'll change the rest}} ok I did it!
    \label{tab:semcor_statistics}
\end{table}

%%%%%%%%%%%%%%%%%%%%%%%%%%%%%%%%%%%%%%%%%%%%%%%%%%%%%%%%%%%

\section{Proportion of each POS in SemCor 3.0}
\label{app:pos_proportion}
In table \ref{tab:pos_proportion_semcor}, we report the percentage of each part of speech in SemCor 3.0 dataset. These values were used to compute the $_{w}$AVG metric reported in Table \ref{tab:semcor_baseline_results}.

\begin{table}[H]
    \centering
    \begin{tabular}{cc}
        \hline
        \textbf{POS} & {\%} \\
        \hline	
        Noun & 0.49 \\
        Adjective & 0.22 \\
        Verb & 0.30 \\
        \hline
    \end{tabular}
    \caption{Percentage of each POS in SemCor Brown1 and Brown2 from which {\semcorwsi} was composed.}
    \label{tab:pos_proportion_semcor}
\end{table}

%%%%%%%%%%%%%%%%%%%%%%%%%%%%%%%%%%%%%%%%%%%%%%%%%%%%%%%%%%%

\section{Best layer for each model in Tables \ref{tab:semcor_baseline_results}, \ref{tab:augmentation_BERT_large_uncased} and \ref{tab:augmentation_BERT_large_ft}}
\label{app:best_layer}
In Tables \ref{tab:best_layer_baseline_models}, \ref{tab:best_layer_augmentation_BERT_large_uncased} and \ref{tab:best_layer_augmentation_BERT_ft}, we provide the results of tuning the \textit{layer} hyperparameter for each model tested.

\begin{table}[H]
    \centering
    \begin{adjustbox}{max width=0.5\textwidth}
    \begin{tabular}{ccccc}
    \textbf{Model} & \textbf{ALL} & \textbf{Verb} & \textbf{Adj} & \textbf{Noun} \\ 
    \hline
    \multicolumn{5}{c}{\textbf{\agsilh}} \\
        BERT-l-Wikt & 23 & 23 & 24 & 23  \\
        BERT-b-u & 11 & 10 & 11 & 11  \\
        MirrorWiC-base & 9 & 9 & 9 & 9  \\
        BERT-l-u & 20 & 17 & 19 & 22  \\
        %ModernBERT-base & 22 & 8 & 20 & 22  \\
        %ModernBERT-large & 28 & 12 & 23 & 27  \\
        %BERT-large-Wikt-POS & NA & 22 & 24 & 18 \\
    \multicolumn{5}{c}{\textbf{X-Means}} \\
        BERT-l-Wikt & 21 & 21 & 23 & 23  \\
        BERT-b-u & 12 & 12 & 12 & 12  \\
        MirrorWiC-base & 12 & 11 & 12 & 11  \\
        BERT-l-u & 21 & 24 & 3 & 24  \\
        %ModernBERT-base & 6 & 16 & 16 & 15  \\
        %ModernBERT-large & 16 & 15 & 6 & 16  \\
        %BERT-large-Wikt-POS & NA & 23 & 23 & 21 \\
    \end{tabular}
    \end{adjustbox}
    \caption{for Table \ref{tab:semcor_baseline_results},  best layer for each PLM on \agsilh and X-means.}
    \label{tab:best_layer_baseline_models}
\end{table}

\begin{table}[h]
    \centering
    \begin{tabular}{l|c|cc}
        \hline
        \textbf{Aug} & \textbf{Base} & \multicolumn{2}{c}{\textbf{Must-link}} \\
         & \textbf{AG$_s$} & \textbf{AG$_{wikt}$} & \textbf{AG$_{s}$}  \\
        \hline
        No & 20 & NA & NA \\
        Wiktionary & 20 & 17 & 20 \\
        Llama 3.1 8B 4bit & 24 & 22 & 17 \\
        GPT-4o & 24 & 19 & 24  \\
        WB (10 per l.) & 16 & 24 & 22  \\
        WB (50 per l.) & 16 & 23 &  16 \\
        WB (100 per l.) & 19 & 20 &  19 \\
        WB (150 per l.) & 20 & 20 & 18  \\
        \hline
    \end{tabular}
    \caption{For Table \ref{tab:augmentation_BERT_large_uncased}, best layer for each data augmentation type on BERT-l-u + AG.}
    \label{tab:best_layer_augmentation_BERT_large_uncased}
\end{table}

\begin{table}[h]
    \centering
    \begin{tabular}{l|c|cc}
        \hline
        \textbf{Aug} & \textbf{Base} & \multicolumn{2}{c}{\textbf{Must-link}} \\
         & \textbf{AG$_s$} & \textbf{AG$_{wikt}$} & \textbf{AG$_{s}$}  \\
        \hline
        No & 23 & NA & NA \\
        Wiktionary & 23 & 23 & 23   \\
        Llama 3.1 8B 4bit & 24 & 23 &  22 \\
        GPT-4o & 24 & 23 & 23  \\
        WB (10 per l.) & 23 & 22 & 20  \\
        WB (50 per l.) & 24 & 22 &  20 \\
        WB (100 per l.) & 22 & 22 & 22  \\
        WB (150 per l.) & 20 & 22 &  20 \\
        \hline
    \end{tabular}
    \caption{For Table \ref{tab:augmentation_BERT_large_ft}, best layer for each data augmentation type on BERT-l-Wikt + AG.}
    \label{tab:best_layer_augmentation_BERT_ft}
\end{table}

\begin{comment}
\begin{table}[h]
    \centering
    \begin{tabular}{l|c|cc}
        \hline
        \textbf{Aug} & \textbf{Base} & \multicolumn{2}{c}{\textbf{Must-link}} \\
         & \textbf{AG$_s$} & \textbf{AG$_{wikt}$} & \textbf{AG$_{s}$}  \\
        \hline
        No & 9 & NA & NA  \\
        Wiktionary & 9 & 11 &  9 \\
        Llama 3.1 8B 4bit & 9 & 9 & 9  \\
        GPT-4o & 9 & 10 & 9  \\
        WB (10 per l.) & 9 & 11 &  9 \\
        WB (50 per l.) & 11 & 11 & 9  \\
        WB (100 per l.) & 10 & 11 & 9  \\
        WB (150 per l.) & 11 & 11 &  9 \\
        \hline
    \end{tabular}
    \caption{For Table \ref{tab:augmentation_MirrorWiC}, best layer for each data augmentation type on MirrorWiC-base + AG.}
    \label{tab:best_layer_augmentation_MirrorWiC}
\end{table}
\end{comment}

%%%%%%%%%%%%%%%%%%%%%%%%%%%%%%%%%%%%%%%%%%%%%%%%%%%%%%%%%%%

\section{Statistical significance: bootstrapping}
\label{app:bootstrapping}
 %\notes{TODO: cite the bootstrapping technique}
 % Dror et al's book is a good reference to cite https://www.amazon.fr/Statistical-Significance-Testing-Language-Processing/dp/3031010469

To evaluate the statistical significance of {\fbcubed} results differences, we apply the bootstrapping test \citep{Dror2020}. Being computationally intensive, we only computed statistical significance for all pairs of systems of Table \ref{tab:semcor_baseline_results} in combination with {\agsilh} as X-means would require rerunning each run 5 times due to its non-deterministic nature. For each pair of models, we verify the null hypothesis that the results difference between two models is due to chance. We sample with replacement the development set of {\semcorwsi} 1000 times and perform clustering on each sample using both models. Then, for each sample, we compute the difference between two models' performance ($\Delta_{sample}$) and check if it is higher than twice the original difference between the 2 models ($\Delta_{obs}$). The p-value is the proportion of samples for which $\Delta_{sample} \leq 2\times\Delta_{obs}$. We reject the null hypothesis when p-value is less than 0.05. In Figure \ref{fig:bootstrapping_for_baselines_ag_all_models}, we present the histograms of bootstrap results for each PLM in combination with \agsilh.

\begin{comment}
\begin{table}[H]
    \centering
    \begin{tabular}{ccc}
    \hline
    pos & $\Delta_{obs}$ & p-value \\ 
    \hline
    all & 0.7 & 1.0 \\
    noun & 0.1 & 0.96 \\
    adj & -0.2 & 0.0 \\
    verb & 2.0 & 0.949 \\
    \hline
    \end{tabular}
    \caption{\todo{if we keep more models, remake the table or remove it} Bootstrapping significance test results comparing \agsilh in combination with two models: BERT-large-Wikt-all+\agsilh and BERT-large-uncased}
    \label{tab:my_label}
\end{table}
\end{comment}

\begin{figure*}
    \centering
    \includegraphics[width=0.9\linewidth]{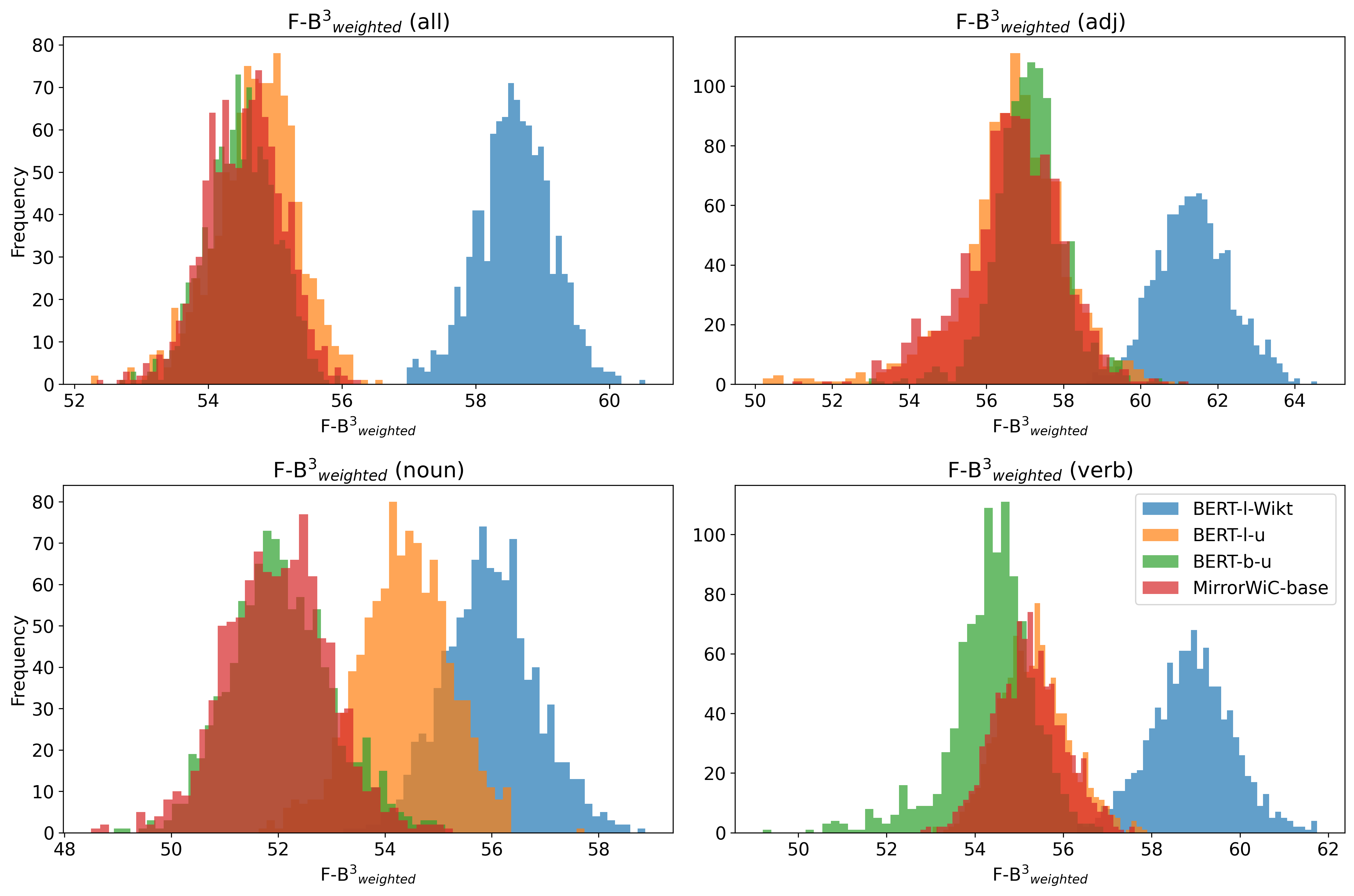}
    \caption{Bootstrapping distribution for 1000 runs using {\agsilh} in combination with all PLMs.}
    \label{fig:bootstrapping_for_baselines_ag_all_models}
\end{figure*}
    
%%%%%%%%
\section{Statistics of added examples from each source}
\label{app:stats_added_examples}

The total number of examples generated or gathered from WikiBooks is presented in Table \ref{tab:statistics_added_examples}. For WikiBooks, we note that the number of examples is not equal to the number of lemmas $\times$ the number of additional examples, as some lemmas are missing from the corpus and some had less examples than required. For LLMs, the number is not equal to the number of instances of \semcorwsi{} $\times$ the number of generated examples, as both models occasionally generated more or less than 3 examples, or refused to perform the task at all.

\begin{table}[H]
    \centering
    \begin{adjustbox}{max width=0.5\textwidth}
    \begin{tabular}{cccccc}
\hline
\textbf{Source} & \textbf{Selection} & \textbf{Verb} & \textbf{Noun} & \textbf{Adj} & \textbf{Total} \\
\hline
Wikt & all per L & 2705 & 4016 & 1805 & 8526 \\
\hline
Llama & 3 per Inst. & 15171 & 16439 & 14852 & 46462 \\
GPT-4o & 3 per Inst. & 14970 & 16154 & 14700 & 45824 \\
\hline
WB & 10 per L & 3511 & 4184 & 3539 & 11234 \\
%wikibooks & 30 per L & 10322 & 11976 & 9672 & 31970 \\
WB & 50 per L & 16910 & 19341 & 15246 & 51497 \\
WB & 100 per L & 32457 & 36421 & 27606 & 96484 \\
WB & 150 per L & 46671 & 51994 & 38756 & 137421 \\
\hline
    \end{tabular}
    \end{adjustbox}
    \caption{Total number of examples added for each POS using different sources. \textbf{per l} = N examples added for each lemma, \textbf{per inst} = N examples added for each instance.}
    \label{tab:statistics_added_examples}
\end{table}

\begin{comment}
\section{Results on \semcorwsi{} test set for all POS}
\label{app:semcor_test_results}
In Table \ref{tab:semcor_test_results}, we provide the results of the current state-of-the-art system for the previous WSI tasks (PolyLM-large), our best unsupervised system (BERT-l-u-AG$_s$+150WB) and our best Wiktionary-using system (BERT-l-Wikt-MustLink-AG$_{wikt}$+100WB), as well as 1 cluster per lemma baseline.

\begin{table}[H]
    \centering
    \begin{tabular}{c|c}
    \hline    
    \textbf{Model} & \textbf{\fbcubed} \\
    \hline
    PolyLM-large & 72.7 \\ 
    BERT-l-uncased+AG$_{s}$+150WB & 74.0 \\
    BERT-l-Wikt+AG$_{wikt}$+ML+100WB & \textbf{76.0} \\
    \hline
    1cpl & 74.4 \\
    \hline
        
    \end{tabular}
    \caption{{\fbcubed} results on \semcorwsi{} test set for all POS of the best previous system (cf. Table \ref{tab:results_on_existing_datasets}) and the best unsupervised and supervised models from Tables \ref{tab:augmentation_BERT_large_uncased} and \ref{tab:augmentation_BERT_large_ft} (for these models we reuse the best layer tuned on the development set).}
    \label{tab:semcor_test_results}
\end{table}
\end{comment}

\section{LLMs details and prompts for direct WSI prompting}
\label{app:prompts}

We tested 3 large language models: the proprietary GPT 4-o (gpt-4o-2024-08-06) \citep{openai2024gpt4technicalreport} and two open-source models: Llama 3.1 8B Instruct\footnote{Version released on July 23, 2024, {\scriptsize \url{https://huggingface.co/unsloth/Meta-Llama-3.1-8B-Instruct} }}\cite{grattafiori2024llama3herdmodels}  and Llama 3.3 70B Instruct (4 bit)\footnote{Version released on Dec 6, 2024, {\scriptsize \url{https://huggingface.co/unsloth/Llama-3.3-70B-Instruct}}. For all Llama models, we use unsloth \citep{unsloth} for faster computation.}.

For the direct WSI prompting, we tested three prompt strategies, where the LLM was asked either to: 1) provide a Python list with cluster numbers for each instance, 2) arrange instance identifiers into Python lists considered as clusters, 3) or assign a sense identifier for each sentence with its index. The last approach yielded the best results, thus, we provide the corresponding prompts below and the results for this approach only.

For SE13, we tuned the prompt using its trial set. For SE10 and {\semcorwsi}, the prompt was tuned using the {\semcorwsi} development set, as the task for both datasets is to predict a single sense per instance. We set the maximum sequence length to 40,000 (to handle lemmas with 500+ instances), the maximum number of new tokens to 4,000, and the default values for the remaining hyperparameters. Model responses were parsed using regular expressions, and missing values were assigned a uniform dummy sense identifier.

\subsection{SemEval 2010 Task 14  and SemCor prompt:}

{\ttfamily
Given the following examples of sentences using the lemma '[LEMMA]', identify the sense of the target lemma for each sentence.

    Examples: 
    
    ----- 
    
    [[INDEX]. [SENTENCE]] 
    
    ----- 
    
For each sentence, your response should be in the format: '[sentence\_index]. [sense\_identifer]'.

Please respond with one sense for each sentence. Please provide answers for all examples. Do not write any explanations. Do not write the sentence in your answer. Only give the sentence index and sense identifier.
}

\subsection{SemEval 2013 Task 11 prompt:}

{\ttfamily
Given the following examples of sentences using the lemma '[LEMMA]', identify the possible senses of the lemma and their level of applicability for each sentence. For each sentence, list the possible senses of the lemma with their corresponding level of applicability (from 0 to 1). \\

    Examples: 
    
    ----- 
    
    [[INDEX]. [SENTENCE]] 
    
    ----- 
    
    For each sentence, your response should be in the format: '[sentence\_index]. [sense\_1/applicability\_1] [sense\_2/applicability\_2']'.
    For example, the answer might look like "100. sense\_1/2", "100. sense\_1/0.8 sense\_2/0.4" or "100. sense\_1/1 sense\_2/0.4 sense\_3/4".

    Please respond with the possible senses of the lemma and their level of applicability for each sentence. Do not write any explanations. Do not write the sentence in your answer. Only give the sentence index, sense identifiers and their level of applicability.
}

%%%%%%%%%%%%%%%%%%%%%%%%%%%%%%%%%%%%%%%%%%%%%%%%%%%%%%%%%%%%

\section{Prompt for generating unlabeled new examples}
\label{app:prompt_generating_examples}
{\ttfamily
Create 3 examples with the target lemma '[LEMMA]' where this lemma is used in the same sense as in the sentence '[SENTENCE]'. 
Separate each example by \textbackslash n and do not give any explanations.
}
%%%%%%%%%%%%%%%%%%%%%%%%%%%%%%%%%%%%%%%%%%%%%%%%%%%%%%%%%%%%

\section{Hyperparameters for clustering experiments of \S \ref{subsec:semcor_baselines_results}}
\label{appendix:hyper_clustering}

We used the scikit-learn implementation of Agglomerative Clustering and silhouette score \citep{scikit-learn}, with following hyperparameters: average linkage with euclidean distance, minimum number of clusters for silhouette score is 2, maximum is 15. More precisely, for a lemma having $n$ instances, silhouette is only defined for numbers of clusters $c$ such as $2 \leq c \leq n-1$. So if $n \geq 3$, we select the number of clusters $c^{*} = min(15, \text{argmax}_{2 \leq c \leq n-1} silh(c)$. If $n = 2$, we return a single cluster. 

For X-means, we used the pyclustering implementation \cite{Novikov2019}, with following hyperparameters: minimum number of clusters is 1, maximum is 15, tolerance is 0.003. %Since X-Means is based on K-means with K++ initialization, it is not deterministic. Therefore, we report the average and standard deviation of 5 runs for each score. 

%%%%%%%%%%%%%%%%%%%%%%%%%%%%%%%%%%%%%%%%%%%%%%%%%%%%%%%%%%%%

\section{Table \ref{tab:results_on_existing_datasets}: results for each POS}
\label{app:perposresults}
In Table \ref{tab:results_on_existing_dataset}, we detail the performance of each model from Table \ref{tab:results_on_existing_datasets} for each part of speech. We note that the results for adjectives from SE10 are absent, as they were not included in the test set.

\begin{table*}[!ht]
    \centering
    \begin{adjustbox}{max width=\textwidth}
    \begin{tabular}{c|cc|cccc|cc}
        \hline
        \textbf{Model} & \multicolumn{2}{c|}{\textbf{SemEval 2013}} & \multicolumn{4}{c|}{\textbf{SemEval 2010}} & \multicolumn{2}{c}{\textbf{\semcorwsi}} \\
        & \textbf{Fuzzy-NMI} & \textbf{Fuzzy-{\fbcubed}} & \textbf{V-M} & \textbf{Paired F-S} & \textbf{NMI} & \textbf{{\fbcubed}} & \textbf{{\fbcubed}} & \textbf{NMI} \\
        \hline
        \multicolumn{9}{c}{\textbf{Verb}} \\
        \hline 
        PolyLM large & \textbf{25.6} &\textbf{67.8} & 45.2 &\textbf{75.6} & 4.5 & 58.7 & 68.5 & 28.9 \\
        PolyLM base & 25.2 & 66.5 & \textbf{45.3} & 71.9 & 4.3 & 54.2 & 65.8 & 27 \\
        LSDP & 18.5[$\pm$0.6] & 59.1[$\pm$0.8] & 43.2[$\pm$1.1] & 66.0[$\pm$0.8] & 4.2[$\pm$0.2] & 60.2[$\pm$0.3] & 65.2[$\pm$0.3] & 23.6[$\pm$0.7] \\
        \hline
        Llama 8B & 2.3[$\pm$0.5] & 57.7[$\pm$0.5] & 13.2[$\pm$0.7] & 53.1[$\pm$1.9] & 6.0[$\pm$0.3] & 54.5[$\pm$1.3] & 57.8[$\pm$1.8] & 19.1[$\pm$1.0] \\
        Llama 70B & 7.1[$\pm$0.6] & 41.2[$\pm$2.3] & 23.3[$\pm$0.7] & 59.5[$\pm$3.1] & 5.6[$\pm$0.3] & 57.2[$\pm$2.6] & \textbf{68.7[$\pm$0.3]} & \textbf{34.2[$\pm$0.6]} \\
        GPT-4o & 15.0[$\pm$1.1] & 57.3[$\pm$2.3] & 33.8[$\pm$2.7] & 67.6[$\pm$4.9] & 5.0[$\pm$0.4] & 51.0[$\pm$4.8] & 63.1[$\pm$1.7] & 26.1[$\pm$0.7] \\
        \hline
        1cpl & 0 & 61.5 & 0 & 72.7 & 0 & \textbf{73.4} & 65.7 & 14 \\
        1cpex & 7.1 & 0 & 25.6 & 0.1 & \textbf{15.7} & 8.2 & 25.5 & 26.9 \\
        \hline
        \multicolumn{9}{c}{\textbf{Noun}} \\
        \hline
        \textbf{Model} & \multicolumn{2}{c|}{\textbf{SemEval 2013}} & \multicolumn{4}{c|}{\textbf{SemEval 2010}} & \multicolumn{2}{c}{\textbf{\semcorwsi}} \\
        & \textbf{Fuzzy-NMI} & \textbf{Fuzzy-{\fbcubed}} & \textbf{V-M} & \textbf{Paired F-S} & \textbf{NMI} & \textbf{{\fbcubed}} & \textbf{{\fbcubed}} & \textbf{NMI} \\
        \hline
        PolyLM large & \textbf{23.4}& \textbf{64.5}& 42.5 & 62 & 7.3 & 42.7 & 74.3 & \textbf{41.9}\\
        PolyLM base & 20.5 & 62.5 & 39.3 & 62.7 & 7.5 & 45.6 & 72.9 & 38.2 \\
        LSDP & 22.2[$\pm$0.6] & 64.3[$\pm$0.5] & \textbf{47.1}[$\pm$1.1]& \textbf{67.4}[$\pm$0.6]& 4.8[$\pm$0.2] & 47.8[$\pm$0.4] & 72.3[$\pm$0.7] & 36.5[$\pm$1.1] \\
        \hline
        Llama 8B & 2.2[$\pm$0.5] & 58.1[$\pm$1.5] & 18.7[$\pm$1.3] & 46.6[$\pm$1.7] & 8.1[$\pm$0.7] & 46.3[$\pm$2.0] & 60.0[$\pm$1.0] & 22.7[$\pm$0.9] \\
        Llama 70B & 10.8[$\pm$0.4] & 46.2[$\pm$2.1] & 33.6[$\pm$1.2] & 42.9[$\pm$6.2] & 9.8[$\pm$0.8] & 44.3[$\pm$1.1] & 65.5[$\pm$2.4] & 23.8[$\pm$3.9] \\
        GPT-4o & 18.3[$\pm$1.2] & 59.9[$\pm$1.3] & 38.1[$\pm$2.0] & 61.3[$\pm$1.5] & 8.5[$\pm$0.4] & 45.4[$\pm$0.7] & 71.4[$\pm$0.6] & 37.3[$\pm$1.1] \\
        \hline
        1cpl & 0 & 61.8 & 0 & 57 & 0 & \textbf{57.6}& \textbf{75.2}& 30.4 \\
        1cpex & 7.1 & 0 & 35.8 & 0.1 & \textbf{22.1}& 7.9 & 24.1 & 19.1 \\
        \hline
        \multicolumn{9}{c}{\textbf{Adjective}} \\
        \hline 
        \textbf{Model} & \multicolumn{2}{c|}{\textbf{SemEval 2013}} & \multicolumn{4}{c|}{\textbf{SemEval 2010}} & \multicolumn{2}{c}{\textbf{\semcorwsi}} \\
        & \textbf{Fuzzy-NMI} & \textbf{Fuzzy-{\fbcubed}} & \textbf{V-M} & \textbf{Paired F-S} & \textbf{NMI} & \textbf{{\fbcubed}} & \textbf{{\fbcubed}} & \textbf{NMI} \\
        \hline
        PolyLM large & 20.7 & 68 & NA & NA & NA & NA & 76& 29.1 \\
        PolyLM base & 23.8 & \textbf{68.7}& NA & NA & NA & NA & 74.9 & 27.1 \\
        LSDP & \textbf{24.4}[$\pm$1.3]& 62.4[$\pm$0.7] & NA & NA & NA & NA & 75.5[$\pm$0.4] & 35.8[$\pm$1.1] \\
        \hline
        Llama 8B & 2.6[$\pm$0.6] & 53.6[$\pm$1.7] & NA & NA & NA & NA & 61.4[$\pm$1.8] & 15.9[$\pm$1.7] \\
        Llama 70B & 8.6[$\pm$0.7] & 45.5[$\pm$2.8] & NA & NA & NA & NA & 58.0[$\pm$2.2] & 24.9[$\pm$0.5] \\
        GPT-4o & 18.0[$\pm$1.8] & 58.1[$\pm$4.2] & NA & NA & NA & NA & 65.7[$\pm$0.9] & 23.5[$\pm$5.3] \\
        \hline
        1cpl & 0 & 59.4 & NA & NA & NA & NA & \textbf{80}& \textbf{39.8}\\
        1cpex & 6.6 & 0 & NA & NA & NA & NA & 22.6 & 16.2 \\
        \hline
    \end{tabular}
    \end{adjustbox}
    \caption{Extension of Table \ref{tab:results_on_existing_datasets} for each part of speech subset.}
    \label{tab:results_on_existing_dataset}
\end{table*}

\begin{comment}
    
\section{Data augmentation (Section \ref{sec:augmentation}): additional results}
\label{app:data_augmentation}
\begin{table}[H]
    \centering
    \begin{tabular}{l|c|cc}
        \hline
        \textbf{Aug} & \textbf{Base} & \multicolumn{2}{c}{\textbf{Must-link}} \\
         & \textbf{AG$_s$} & \textbf{AG$_{wikt}$} & \textbf{AG$_{s}$} \\
        \hline
No & 70.2 & NA & NA \\
Wiktionary & 71.1 & 71.3 & 70.8 \\
GPT-4o & 70.5 & 73.2 & 72.7 \\
Llama 3.1 8B 4bit & 70.6 & 73.2 & 72.9 \\
WB (10 per l.) & 72. & 	73.1 & 73.1 \\
WB (50 per l.) & 73.4 & 73.9 & 73.4 \\
WB (100 per l.) & 73.5 & 74.3 & 73.7 \\
WB (150 per l.) & \textbf{73.7} & \textbf{74.5} & \textbf{73.9} \\
 \hline
    \end{tabular}
    \caption{Comparison of augmentation methods for Mirror-WiC-base (self-supervised)}
    \label{tab:augmentation_MirrorWiC}
\end{table}
\end{comment}

%%%%%%%%%%%%%%%%%%%%%%%%%%%%%%%%%%%%%%%%%%%%%%%%%%%%%%%%%%%%

\section{Datasets and LLMs Licenses}

In our experiments, we use WikiBooks part of the BigScience corpus distributed under the BigScience RAIL License available at: \url{https://huggingface.co/spaces/bigscience/license}. Additionally, we use the Wiktionary DBnary dataset, released under the Creative Commons Attribution-ShareAlike 3.0 license. We introduce a new evaluation framework (\semcorwsi) based on SemCor 3.0, which is the property of Princeton University. The corresponding license is included within the SemCor package, accessible at \url{http://web.eecs.umich.edu/~mihalcea/downloads/semcor/semcor3.0.tar.gz}.

Considering Llama models, we use Llama 3.1 (license available at \url{https://www.llama.com/llama3_1/license/}) and Llama 3.3 models (license available at \url{https://www.llama.com/llama3_3/license/}).

\end{document}